\documentclass{article}

\usepackage{arxiv}

\usepackage{newtxtext}
\usepackage{newtxmath}

\usepackage[utf8]{inputenc} % allow utf-8 input
\usepackage[T1]{fontenc}    % support for Cyrillic (T2A) and Latin (T1)
\usepackage[russian,english]{babel} % hyphenation rules for Russian and English
\usepackage{hyperref}       % hyperlinks
\usepackage{url}            % simple URL typesetting
\usepackage{booktabs}       % professional-quality tables
\usepackage{amsfonts}       % blackboard math symbols
\usepackage{nicefrac}       % compact symbols for 1/2, etc.
\usepackage{microtype}      % microtypography
\usepackage{cleveref}       % smart cross-referencing
\usepackage{lipsum}         % Can be removed after putting your text content
\usepackage{graphicx}
\usepackage{natbib}
\usepackage{doi}
\usepackage{authblk}

\usepackage{xcolor}
\definecolor{jsonstring}{rgb}{0.16,0.45,0.12}
\definecolor{jsonkey}{rgb}{0.55,0.06,0.55}
\definecolor{jsonnumber}{rgb}{0.75,0.15,0.15}

\title{Parameter-Efficient Adaptation of Large Language Models to a Low-Resource Agglutinative Language: A Corpus, Datasets, and a Reproducible Benchmark for the Bashkir Language}

\newbox{\orcid}\sbox{\orcid}{\includegraphics[scale=0.06]{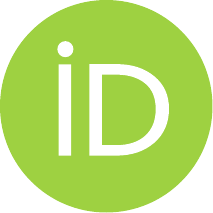}}

\author[1]{%
    \href{https://orcid.org/0000-0003-2525-1183}{\usebox{\orcid}\hspace{1mm}M. K. Arabov\thanks{\texttt{MKArabov@kpfu.ru}}}%
}
\author[1]{S. S. Khaybullina\thanks{\texttt{khaybulinas@mail.ru}}}
\affil[1]{Department of Data Analysis and Programming Technologies, Institute of Computational Mathematics and Information Technologies, Kazan (Volga Region) Federal University, Kazan, Russia}

\renewcommand{\shorttitle}{Parameter-Efficient Adaptation of LLMs to Bashkir}

\hypersetup{
    pdftitle={Parameter-Efficient Adaptation of Large Language Models to a Low-Resource Agglutinative Language: A Corpus, Datasets, and a Reproducible Benchmark for the Bashkir Language},
    pdfsubject={cs.CL},
    pdfauthor={M. K. Arabov and S. S. Khaybullina},
    pdfkeywords={Bashkir language, low-resource languages, agglutinative languages, QLoRA, continual pretraining, tokenisation, LLM-as-Judge, reproducible benchmark},
}

\begin{document}
\maketitle
\begin{abstract}

The Bashkir language belongs to low-resource agglutinative languages with Cyrillic script and remains poorly represented in the ecosystem of modern large language models. This work presents an open resource package for the Bashkir language and a systematic benchmark of large language model adaptation methods: the Bashkir Web Corpus (71\,567 documents, 46.9 million tokens, 16 sources) and three labelled datasets for text classification: binary (16\,994 documents), multi-class (17\,897 documents, 19 categories) and multi-label (22\,318 documents, 14 labels). The classification benchmark was conducted on the first two datasets; the multi-label dataset is published as a resource and is not used in the present benchmark. Experiments were conducted on eight architectures in the range 82M--8B parameters (DistilGPT2, GPT-2 base, GPT-2 medium, Phi-3-mini, Mistral-7B, YandexGPT-5-Lite-8B, mBERT, XLM-R) and three groups of adaptation strategies: full fine-tuning, QLoRA and in-context learning (zero-shot and few-shot); in addition, the benefit of continual pretraining (CPT) on the Bashkir corpus was investigated.

Four results constitute the main contribution of the work. First, QLoRA adaptation of the 8B model YandexGPT-5-Lite-8B on binary classification reaches macro-F1 $0.736$--$0.748$ (at ranks $r=8$ and $r=16$) versus $0.653$ for XLM-R (Full~FT, $278$M) and $0.650$ for mBERT (Full~FT, $178$M); in a paired comparison of accuracy of QLoRA configurations without CPT ($0.737$ for YandexGPT-5-Lite-8B versus $0.659$ for XLM-R and $0.652$ for mBERT), the difference is significant after Holm correction for $15$ comparisons ($p_{\text{Holm}} < 0.001$). This comparison simultaneously changes the adaptation method, the size and the architecture of the model; it is interpreted as an empirical observation for the selected configurations, and not as a pure comparison of methods, and a comparison of QLoRA with full fine-tuning on the same architectures was not conducted in the present work. Second, YandexGPT-5-Lite-8B significantly outperforms Mistral-7B and Phi-3-mini in accuracy on multi-class classification ($0.606$ versus $0.547$ and $0.506$; $p_{\text{Holm}} < 0.001$; point estimates of macro-F1 are $0.368$ versus $0.282$--$0.298$ for Mistral-7B and $0.119$--$0.166$ for Phi-3-mini; a statistical test on macro-F1 was not conducted) and in BPB ($0.462$ versus $0.549$ and $0.704$; $p_{\text{Holm}} < 0.001$). According to native-speaker evaluations, YandexGPT-5-Lite-8B significantly outperforms Mistral-7B on all three criteria ($3.28$ versus $2.52$, $p_{\text{Holm}} < 0.05$, $n=11$); the difference with Phi-3-mini ($2.22$) in the current sample ($n=11$ and $n=9$) does not reach the significance level because of the small sample size (Section~\ref{s6:human}). Zero-shot ICL shows a pointwise advantage ($0.58$--$0.68$ versus $0.36$--$0.51$ macro-F1), but a formal test was not conducted for it; on binary classification, the pointwise advantage of YandexGPT-5-Lite-8B is not statistically confirmed within the QLoRA group ($p_{\text{Holm}} \geq 0.084$). Third, the effect of CPT depends both on the base model and on the task: in terms of perplexity, CPT is significant for all three models, but with opposite sign --- it improves for Phi-3-mini ($\Delta = +0.58$) and YandexGPT-5-Lite-8B ($\Delta = +0.33$), and worsens for Mistral-7B ($\Delta = -1.15$, $p_{\text{Holm}} < 0.001$); on binary classification, by contrast, CPT gives the largest gain precisely for Mistral-7B ($+4.0$ points of macro-F1 with worsening perplexity). Fourth, the fertility rate of the tokeniser on Bashkir varies by a factor of $2.67$ (from $2.92$ to $7.80$) and is consistent with adaptation quality on a sample of five models, whereas perplexity under different tokenisers is unsuitable for cross-model comparison.

Practical recommendations: under a limited computation budget, preference should be given to models with a Cyrillic tokeniser and Russian-language pretraining; for evaluating generation quality in a low-resource language, direct scoring by LLM judges reproduces native-speaker evaluations for grammar and coherence ($\rho = 0.33$--$0.46$), but not for language correctness ($\rho = 0.16$, $p = 0.11$); native-speaker evaluation remains necessary.
\end{abstract}


\section{Introduction}\label{s1}

Bashkir belongs to the Turkic language family and is one of the state languages of the Republic of Bashkortostan. Despite this, Bashkir has an extremely limited digital presence: the volume of machine-readable texts, annotated corpora and pretrained language models is substantially smaller than for languages with a high digital presence and than for a number of closely related Turkic languages.

Over the past few years, large language models have made a qualitative leap: growth in the number of parameters, the transition to instruction tuning, the emergence of parameter-efficient adaptation methods and open models have made large language models the main instrument of natural language processing \cite{comparative2026lora, burnashev2026parameter}. This development, however, is unevenly distributed across languages. The overwhelming majority of modern models are trained on English, which leads to digital inequality: languages with a limited digital presence --- including Bashkir --- are effectively excluded from the ecosystem of modern language technologies.

This inequality manifests itself at three levels. First, tokenisers of multilingual models fragment agglutinative word forms into excessively long subsequences of subtokens, which reduces processing quality \cite{eyupoglu2019two, greshanin2018network}. Second, the absence of large open corpora limits the possibilities of both pretraining from scratch and continual training \cite{khudiakova2025bashqort}. Third, existing linguistic resources for Bashkir were created within a paradigm oriented towards search and aggregation of data rather than machine learning \cite{sirazitdinov2025machine}.

The digitisation of the Bashkir language began in the 2000s and is associated primarily with the creation of the Machine Fund of the Bashkir Language --- an integrated information system developed at the Institute of History, Language and Literature, Ufa Federal Research Centre, Russian Academy of Sciences (IIL UFRC RAS) since 2006 \cite{sirazitdinov2010bashkir, sirazitdinov2025machine}. The structure of the Fund includes a general word list, lexicographic, grammatical and experimental-phonetic databases, catalogues of manuscript and early printed books, a dialectological database, and corpus databases of prose, journalistic and folklore texts \cite{sirazitdinov2025machine}. The total number of linguistic data databases exceeds 75. These resources became the basis for subsequent corpus and applied developments \cite{sirazitdinov2024applied}.

However, the architecture of the Machine Fund, developed on the basis of the Oracle relational database management system, is oriented towards search and aggregation of linguistic information and is not intended for training neural network models. The resulting gap between accumulated linguistic data and modern machine-learning methods determined the need to create open corpora suitable for pretraining and fine-tuning transformer architectures.

In parallel, a strand emerged in Turkic studies concerned with the application of natural language processing methods to low-resource languages. Studies show that transfer of knowledge from closely related languages, multilingual training and parameter-efficient fine-tuning make it possible to achieve acceptable quality even with limited data volumes \cite{saykhunov2019multilingual, khusainov2021multilingual, mirzakhalov2021large, gatiatullin2024turkic, suleimanov2026methods}. At the same time, until recently there were no systematic studies for Bashkir comparing different strategies for adapting large language models.

The relevance of this work is determined by three factors. First, Bashkir remains a low-resource language, and existing open corpora have a limited volume and source imbalance \cite{khudiakova2025bashqort}. Second, parameter-efficient fine-tuning methods that have proved themselves in other languages \cite{comparative2026lora, burnashev2026parameter, arabov2026combining} have not been systematically tested on Bashkir material. Third, practical applications --- machine translation \cite{karpov2026no, maxim2026script, tyurin2026devlake}, speech recognition and synthesis \cite{radford2022robust, orel2023speech, bekarystankyzy2024multilingual}, text classification \cite{smetanina2020software, arabov2026combining, arabov2026comprehensive} --- require reproducible benchmarks and open datasets, which remain insufficient for Bashkir.

This study aims to fill this gap. The work presents three groups of artefacts and results. The first group is resources: the open Bashkir Web Corpus with 46.9 million tokens, as well as three annotated datasets for binary, multiclass and multilabel classification. The second group is empirical results: the application of QLoRA to medium-scale generative models, an assessment of continual pretraining on the Bashkir corpus as a preliminary stage before downstream adaptation, a comparison of in-context learning strategies (zero-shot and few-shot) with parameter updates, and full fine-tuning of small architectures as a baseline. The third group is analysis: measurement of tokeniser fertility and validation of LLM judges against ratings by native Bashkir speakers. Experiments were conducted on eight large language model architectures in the range from 82M to 8B parameters: three small models of the GPT-2 family (DistilGPT2, GPT-2 base, GPT-2 medium), three medium-scale generative models (Phi-3-mini, Mistral-7B, YandexGPT-5-Lite-8B) and two encoder-only baselines (mBERT, XLM-R). All artefacts --- experiment code, preprocessing scripts, trained adapters and the corpus itself --- are released under open licences to ensure reproducibility.

The study addresses four research questions. \textit{RQ1:} How does QLoRA adaptation of generative models of $3{,}8$--$8$B compare with full fine-tuning of encoder-only baselines (mBERT, XLM-R) on Bashkir text classification tasks? \textit{RQ2:} Does continual pretraining on the Bashkir corpus produce a measurable gain in perplexity and downstream quality relative to QLoRA without CPT, and is the effect the same for different base models? \textit{RQ3:} How is tokeniser fertility related to adaptation quality, and does it explain the gap between models with a Cyrillic vocabulary and a BPE vocabulary? \textit{RQ4:} Do LLM judges reproduce the ratings of native Bashkir speakers, and which generation quality criteria remain inaccessible to automatic evaluation?

The work is organised as follows. Section~\ref{s2} contains a review of the literature on the digitisation of the Bashkir language, corpus linguistics, the morphology of agglutinative languages, PEFT and the application of PEFT to Turkic languages. Section~\ref{s3} describes the corpus, datasets, models, adaptation strategies, hyperparameters and metrics. Section~\ref{s4} presents results on language modelling, continual pretraining, binary and multiclass classification, in-context learning, generation quality and tokeniser analysis. Section~\ref{s5} discusses the factors behind the superiority of YandexGPT-5-Lite-8B, the selective effect of CPT, the divergence between LLM judges and native speakers, the role of tokenisation and practical recommendations. Section~\ref{s6} formulates limitations, and Section~\ref{s7} concludes and outlines directions for further work.

\section{Related Work}\label{s2}

\subsection{Digitisation of the Bashkir Language and the Formation of Linguistic Resources}\label{s2:digitization}

The first works describing the process of digitising the Bashkir language date from the early 2010s. The article \cite{sirazitdinov2010bashkir} describes in detail the experience of creating national font layouts, the integration of the Bashkir language into operating systems together with Microsoft, and the conceptual foundations and launch of the information system ``Machine Fund of the Bashkir Language''. For modern natural language processing, this work is a historical and methodological source explaining the origin and initial architecture of the main linguistic resource.

A generalising article is \cite{khisamitdinova2016bashkir}, which summarises the development of Bashkir philology in the post-Soviet period. Particular attention is paid to the creation of the ten-volume ``Academic Dictionary of the Bashkir Language'' and the conversion of its dictionary databases into digital format --- the creation of electronic glossaries and terminological data banks. These lexicographic developments subsequently became the basis for building morphological analysers and automatic dictionaries.

The most complete description of the structure and functional capabilities of the Machine Fund is presented in \cite{sirazitdinov2025machine}. The author describes the organisation of ten large sections of the Fund (general word list, lexicographic, grammatical, experimental-phonetic databases, catalogues of manuscript and early printed books, dialectological database, corpus databases). For natural language processing tasks, this material is the main source for understanding how the annotation of source data is organised and where aggregated characteristics are extracted from.

A retrospective analysis of scientific and practical activity in the field of applied linguistics at IIL UFRC RAS over more than 30 years is presented in \cite{sirazitdinov2024applied}. Over this period, an infrastructure was created for the functioning of the national language in computer technologies, theoretical foundations for an automatic morphological analyser, and phonetic, lexical and full-text linguistic databases, which made it possible to develop the information systems ``Machine Fund of the Bashkir Language'' and ``Terminological Data Bank''.

\subsection{Corpus Linguistics of the Bashkir Language}\label{s2:corpus}

Practical aspects of creating corpora of the Bashkir language are considered in a number of works. In \cite{ishmukhametova2014fiction}, the experience of creating a subcorpus of literary texts of the Machine Fund is described: issues of data cleaning, tokenisation and formats for encoding metatextual annotation (genre, author, year) are considered. These aspects are fundamentally important for selecting and filtering quality data during pretraining of transformer models.

In \cite{buskunbaeva2018one}, using the material of the folklore corpus, types of one-member sentences in the proverbial fund of the Bashkir language are described: the structural classification of proverbial constructions and the principles of their annotation in the corpus are considered. This material is important for developing syntactic analysers and for understanding how the annotation of folklore texts in Bashkir corpora is organised.

The dialectological aspect of corpus linguistics is presented in \cite{sirazitdinov2016dialect}, which describes an applied component of the Machine Fund --- an integrated Dialectological Database consisting of a lexical subfund, a textological corpus and an atlas. For natural language processing tasks, the article is valuable for its description of the principles of annotation, tagging and computer recognition of regional variations of the Bashkir folklore and living continuum.

Issues of code-switching using the material of the journalistic corpus are considered in \cite{buskunbaeva2022code}. Structural types of switches and borrowings are analysed, which is necessary for correct tokenisation, dataset cleaning and training transformer models to process the Russian--Bashkir bilingual context. This problem is particularly relevant because real Runet texts often contain a mixture of Russian and Bashkir.

The development of an audio corpus of the eastern dialect of the Bashkir language is the subject of \cite{buskunbaeva2020audio}. The methodology of creating an audio corpus, the principles of phonetic and acoustic annotation, the digitisation of live recordings and the alignment of audio segments with text transcripts are described. This material is important for developers of voice assistants, speech recognition systems and speech synthesis models.

Parameters of oral speech segmentation using the material of a media audio corpus are studied in \cite{khamitova2024phrase}: using the ELAN program, syntagms and phrase volume are analysed. In the field of speech-to-text conversion, this publication provides mathematical and linguistic parameters for segmenting the audio stream and configuring acoustic models for Bashkir speech recognition.

Finally, the Bashqort Raw Corpus dataset \cite{khudiakova2025bashqort} is a collection of Bashkir texts intended for continual training of large language models. The resource is published on the Hugging Face platform and is one of the few open corpora of the Bashkir language.

\subsection{Lexicography, Orthography and Formal Grammar}\label{s2:lexicography}

The transition of classical lexicography into the digital environment is considered in \cite{yagafarova2015online}: the structure, strengths and weaknesses of existing electronic and online Bashkir dictionaries are analysed. In solving natural language processing tasks, these materials are used to develop automatic parsers, electronic glossaries and machine translation systems.

A fundamental classification of the principles of nomination and thematic groups of Bashkir vocabulary is presented in the doctoral dissertation \cite{yagafarova2019onomasiological}. In IT developments, this structure is used to design Bashkir WordNet-like networks, build semantic ontologies, knowledge graphs and named entity extraction systems.

The principles of creating the information and reference portal on Bashkir orthography Belemle.ru are described in \cite{yagafarova2025belemle}. For developers, the material is valuable for its description of standardised rules underlying spell-checking algorithms, automatic correction of typos and normalisation of Bashkir text before it is fed into neural network models.

The formalisation of Bashkir syntax from the standpoint of computer modelling is presented in \cite{sirazitdinov2021sentence}: the author identifies strict models and schemes of verbal government using the material of generalised-personal proverbs. In natural language processing tasks, the formalisation of grammar makes it possible to specify rules for accurate semantic-syntactic parsing of text constructions.

\subsection{Applied Platforms and Services}\label{s2:platforms}

The software architecture of the Bashkortsoft ecosystem is considered in \cite{gataullin2022bashkortsoft}: modules for intelligent text analysis, an API for integrating Bashkir dictionaries into third-party IT solutions, algorithms for automatic spell-checking and normalisation of noisy user input are described. This material is important as an example of implementing applied text preprocessing pipelines before feeding data into neural networks.

The process of transferring scientific databases of the Machine Fund into commercial and publicly accessible web interfaces of the Bashkortsoft project is described in \cite{sirazitdinov2021lingvosoft}. Technical problems of matching encodings, parsing large volumes of dictionary entries and optimising search queries for online translators operating in real time are considered.

The concept of collecting open legally clean datasets for training large language models is presented in \cite{gataullin2024digital}: problems of copyright in texts are analysed, and the integration of Bashkir translation on modern IT platforms is described. The architecture of the Bashkorttele repository, including audiobooks, annotated Bashkir epics and works by classical authors, is described in \cite{yusupova2026bashkorttele}. The data alignment format (each audio fragment is strictly aligned with text in Bashkir and Russian) makes this work a methodological guide for training speech-to-text systems.

Scaling of the Bashkorttele data repository is described in \cite{kilsenbaev2026parallel}: the structure of a trilingual parallel spoken corpus of 10,000 phrases is disclosed, as well as the integration of an archive of live Bashkir speech with a volume of 53.3 hours of clean audio based on radio programmes of the State Television and Radio Broadcasting Company ``Bashkortostan''.

\subsection{Morphology of Agglutinative Languages}\label{s2:morphology}

A two-level morphological description of the Bashkir language is presented in \cite{eyupoglu2019two}: the description is based on a lexicon of root words, implemented using XML and connected to the Nuve framework. The phonetic rules of the Bashkir language are encoded using 41 two-level rules. This work is one of the few attempts at a formal description of Bashkir morphology in terms of two-level morphology.

A network model of agglutinative morphology based on graph theory is proposed in \cite{greshanin2018network}. The model is implemented for modern Bashkir and for Old Turkic; on its basis, a morphological parser for the Old Turkic language was created.

A neuro-symbolic approach to the morphology of Bashkir nouns is developed in \cite{isangulov2025neuro, isangulov2026neuro}. In \cite{isangulov2025neuro}, a hybrid method is presented that combines a large language model and a finite-state automaton to ensure morphological correctness during text generation in agglutinative languages: the system automatically extracts rules from corpus data, generates context-dependent regex rules and LEXC morphotactics; the integration of the finite-state automaton into the language-model loop improved machine translation quality from BLEU 16.14 / ChrF 45.13 to BLEU 25.71 / ChrF 50.87 without retraining the translator. In \cite{isangulov2026neuro}, the previous version of the pipeline is extended: 69.60\,\% coverage was achieved and 174 of 250 forms were recognised on the same benchmark.

\subsection{Turkic Languages: Corpora and Machine Translation}\label{s2:turkic_mt}

The principles of building multilingual Turkic corpora are considered in \cite{saykhunov2019multilingual}: the architecture of cross-lingual models for closely related languages is studied, and transfer learning algorithms are proposed to improve the quality of machine translation and text classification in the low-resource Bashkir language.

A multilingual neural machine translation system for seven Turkic--Russian language pairs is presented in \cite{khusainov2021multilingual}: parallel corpora with a total volume of about 7.9 million sentence pairs were created, and for the first time for Turkic languages, transfer learning experiments were conducted on a combined Russian--Turkic parallel corpus.

A large-scale study of machine translation in the Turkic language family is presented in \cite{mirzakhalov2021large}: a parallel corpus covering 22 Turkic languages was created, bilingual baseline models for 26 language pairs were proposed, as well as new test sets in three different translation domains.

The toolkit of the ``Turkic Morpheme'' portal for data augmentation in low-resource machine translation is described in \cite{gatiatullin2024turkic}: an augmentation method based on morphological translation showed an increase in machine translation quality by the BLEU metric, especially effectively for closely related languages with similar grammar and morphology.

Methods for creating machine translation systems for low-resource languages are considered in \cite{suleimanov2026methods}: an approach combining data augmentation and knowledge transfer is proposed; BERT is chosen as the architecture, used both as an encoder and as a decoder. Parallel corpora with a total volume of about 30 million sentence pairs were collected.

The Russian--Bashkir parallel corpus \cite{bak_rus_parallel} contains 9.7 million sentence pairs filtered by semantic similarity. The corpus combines real translations and synthetic data obtained by back-translation.

The application of parameter-efficient fine-tuning to the task of Russian--Bashkir machine translation is presented in \cite{karpov2026no}: fine-tuning of nllb-200-distilled-600M with LoRA on synthetic data achieved chrF++ 46.94 for Bashkir. In \cite{maxim2026script}, adaptation of the Tencent HY-MT1.5-7B model with LoRA is described for correcting output from Arabic script to Cyrillic in Russian--Kazakh and Russian--Bashkir pairs. In \cite{tyurin2026devlake}, a comparative study of NLLB-200, M2M-100 and MarianMT architectures was conducted; the best system --- a fine-tuned NLLB-200-1.3B with exact-match extraction --- reached chrF++ 52.67.

\subsection{Speech Processing for Turkic Languages}\label{s2:speech}

The Whisper speech recognition model \cite{radford2022robust} was adapted to Bashkir by reducing the vocabulary to 16,384 tokens. The Bashkort TTS dataset \cite{bashkorttts} is the largest open dataset for speech synthesis in Bashkir: 62,852 audio files, eight speakers and three speech styles.

Multilingual speech synthesis for Turkic languages using transliteration is presented in \cite{yeshpanov2023multilingual}: a system based on Tacotron~2 was trained only on Kazakh data, and to generate speech in other Turkic languages, letters of the alphabets are mapped to International Phonetic Alphabet symbols, which are then converted into Kazakh letters.

Transfer learning from Kazakh for speech recognition of nine Turkic languages is studied in \cite{orel2023speech}: most models built using transfer learning outperformed monolingual baselines; the most significant improvement was achieved for Yakut. Multilingual end-to-end speech recognition for five agglutinative Turkic languages with a common Cyrillic script is presented in \cite{bekarystankyzy2024multilingual}: multilingual training improved recognition scores for all languages except Bashkir. In \cite{bekarystankyzy2025integrated}, an integrated end-to-end multilingual speech recognition architecture is proposed, using Connectionist Temporal Classification and an attention mechanism with a language model.

\subsection{Sentiment Analysis and Text Classification}\label{s2:sentiment}

A software solution for building classifiers for natural language text analysis is presented in \cite{smetanina2020software}: the task of sentiment analysis of Bashkir news texts is formulated, and a hybrid approach including dictionary-based methods and machine-learning methods is proposed. In \cite{suleimanov2020models}, a formal description of the term ``opinion'', the sentiment analysis task, approaches to its solution and ready-made software implementations are presented; the developed solution based on the support vector machine with stochastic gradient descent demonstrated the highest scores in terms of precision, recall and F-measure.

\subsection{Parameter-Efficient Fine-Tuning}\label{s2:peft}

The methods of parameter-efficient fine-tuning arose as a response to the high cost of full fine-tuning of large language models. LoRA \cite{hu2022lora} freezes pretrained weights and adds trainable low-rank matrices, which reduces the number of trainable parameters by several orders of magnitude. QLoRA \cite{dettmers2023qlora} extends this approach with 4-bit quantisation to the NF4 format, which makes it possible to fine-tune models with billions of parameters on a single GPU. DoRA \cite{liu2024dora} decomposes pretrained weights into magnitude and direction, applying LoRA only to the direction update; this decomposition increases the learning capacity and stability of the method. QDoRA combines DoRA weight decomposition with 4-bit quantisation, which makes the method applicable under limited GPU memory.

In this work, QLoRA is used among the methods listed above: it is implemented in all PEFT experiments and forms the basis of the comparative analysis. Unquantised LoRA does not fit into 24~GB of NVIDIA L4 memory for 7--8B models and was not run in the configuration reported here; DoRA and QDoRA are planned as an extension and require further validation of compatibility with 4-bit quantisation in the Colab environment.

\subsection{Application of PEFT to Turkic Languages}\label{s2:peft_turkic}

A comparative study of LoRA and QLoRA methods for the Bashkir language is presented in \cite{comparative2026lora}: experiments were conducted on a corpus of 71 thousand documents (46.9 million tokens) using DistilGPT2, GPT-2, Phi-2, Qwen2.5-7B, DeepSeek-7B and Mistral-7B. The best perplexity was obtained for GPT-2 medium with full fine-tuning (3.34), while QLoRA for Mistral-7B (3.79) and Phi-2 (3.81) achieved comparable quality with 40 times fewer trainable parameters.

Application of LoRA and QLoRA to the Tatar language is studied in \cite{burnashev2026parameter}: QLoRA adaptation of Mistral-7B reached a perplexity of 3.98 on the test set, comparable to fully fine-tuned gpt2-medium (2.91). Mistral-7B outperformed Qwen2.5-7B (perplexity 5.44) and DeepSeek-7B (perplexity 5.32) in adaptation to Tatar. Increasing the LoRA rank from 8 to 16 consistently improved performance across all architectures.

Combining classical and transformer approaches for text classification and topic modelling of the Tatar language is presented in \cite{arabov2026combining}: on a news corpus of 153,162 documents, classical approaches (TF-IDF+LR, FastText+LR) and transformer architectures (mBERT, XLM-R, DistilBERT) were systematically compared on three tasks --- binary, multiclass (9 categories) and multilabel (13 categories).

Modelling subword embeddings for low-resource languages in the context of Industry~4.0 is presented in \cite{arabov2026modelling}: using Tatar as an example, a comparative analysis of Word2Vec and FastText architectures was carried out; subword approaches using character n-grams significantly improved robustness to rare and unseen word forms.

A comprehensive study of Tatar text classification methods --- from corpus creation and Tatar2Vec vector models to benchmark construction --- is presented in \cite{arabov2026comprehensive}: the TatCorp 222M corpus, TatarTokenizers tokenisers, Tatar2Vec models and an annotated dataset of 13,160 texts in ten topic categories were created. It was established that the use of subword BPE tokenisation in the task of topic categorisation of an agglutinative language systematically reduces classification quality.

\section{Methodology}\label{s3}

\subsection{Overall Research Design}\label{s3:overview}

\begin{figure*}[ht]
\centering
\includegraphics[width=\textwidth]{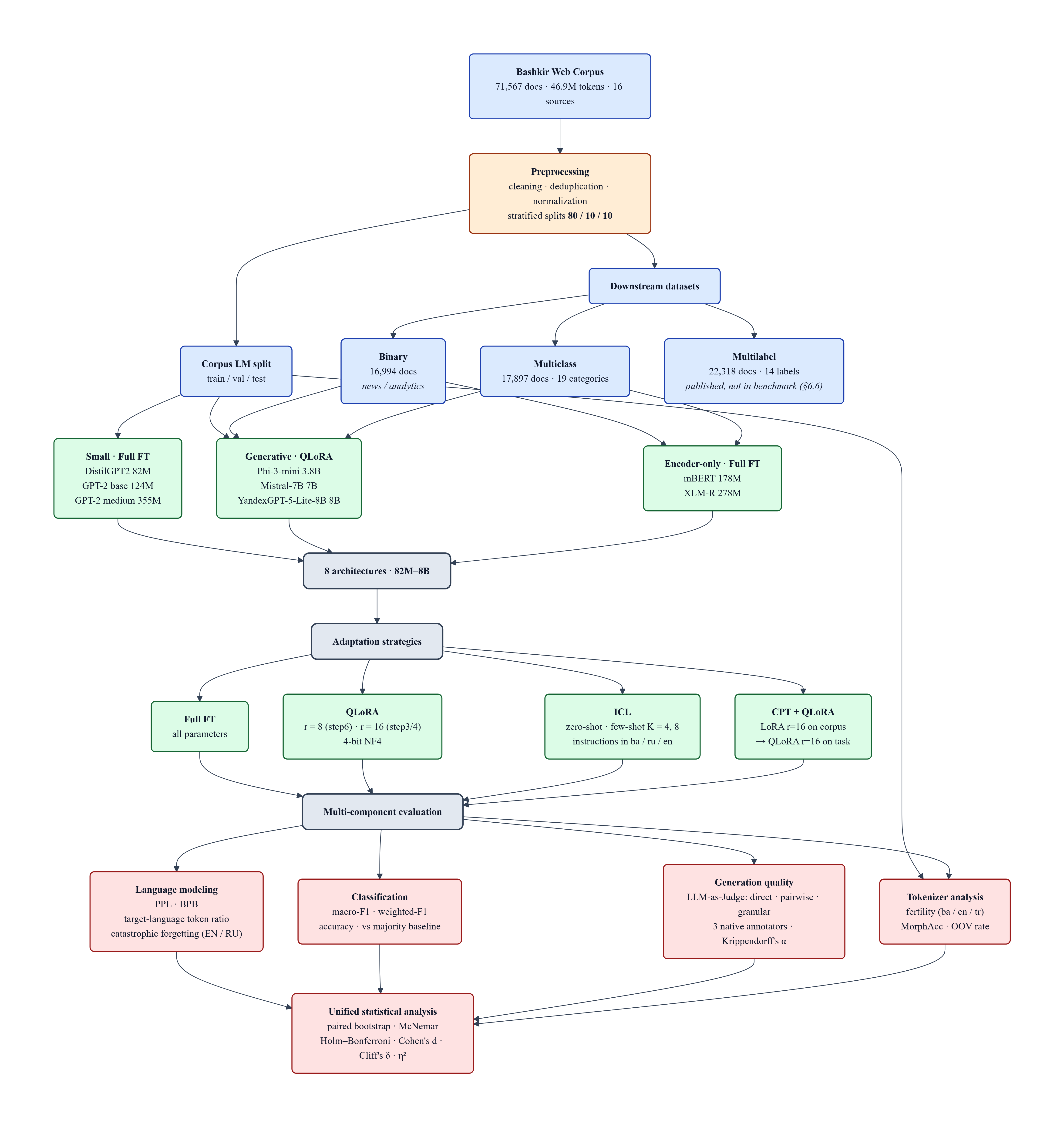}
\caption{Overall experimental pipeline}
\label{fig:pipeline}
\end{figure*}

The experimental study is organised as a six-stage pipeline presented in Figure~\ref{fig:pipeline}. In the first stage, the Bashkir Web Corpus is constructed. In the second, preprocessing and stratified splitting of the data in an 80/10/10 proportion are performed. In the third, three datasets for classification tasks are created from the news sources of the corpus: binary, multiclass and multilabel. In the fourth, eight large language model architectures are selected in the range from 82M to 8B parameters. In the fifth, three groups of adaptation strategies are applied: parameter-efficient fine-tuning, in-context learning and full fine-tuning. In the sixth, a multi-component evaluation is conducted using metrics of language modelling, classification, computational efficiency and generation quality. This organisation makes it possible to trace the path from the source data to the final conclusions sequentially and ensures the reproducibility of each stage.

\subsection{The Bashkir Web Corpus}\label{s3:corpus}

The resource basis of the study is the Bashkir Web Corpus \cite{bashkirnlpworld2026webcorpus} --- a corpus of the Bashkir language with a volume of 71,567 documents and about 46.9 million tokens, collected from 16 open internet resources including news portals, literary journals, social media, books and Wikipedia. The distribution of sources is uneven: the Bashkir Wikipedia (\texttt{bawiki\_cyrillic}) provides 42,262 documents, which constitutes 59.0\% of the corpus, the portal \texttt{azatliqorg} provides 7,541 documents (10.5\%), while the remaining 14 sources are represented by volumes from 151 to 2,139 documents and together cover 30.5\%. A document is described by five fields: \texttt{title} (headline, available for 99.4\% of documents), \texttt{content} (main text), \texttt{category} (topic category, 97.8\%), \texttt{source} (source identifier) and \texttt{text} (the combined string \texttt{title} + \texttt{content}, prepared for input to the model). The corpus is available in two configurations: \texttt{full} (71,567 documents) and \texttt{sample} (1,000 documents for debugging). The licensing model follows the practice of large web corpora such as HPLT and OSCAR: the source texts retain the rights of authors and publishers, while the structured compilation, metadata and original annotations are distributed under the CC BY-SA~4.0 licence with an active notice-and-takedown procedure and a commitment to respond within 14 working days. The main limitation of the corpus is the imbalance of sources: 59\% of documents belong to the encyclopaedic genre, which creates a stylistic shift towards the formal register. Additional limitations are the absence of publication dates, residual web noise and domain unevenness, which must be taken into account when training models and interpreting results.

\subsection{Datasets for Downstream Tasks}\label{s3:datasets}

Three datasets covering text classification tasks of varying complexity were formed on the basis of the news sources of the corpus. The binary classification dataset \cite{arabov2026bashkirbinary} includes 16,994 documents with two balanced classes: news (label~1) and analysis (label~0), with 8,497 examples in each. The labelling was performed automatically on the basis of normalised source categories: all news categories were assigned to label~1, the rest to label~0. Balancing was achieved by undersampling, during which 7,434 analytical articles were removed. The data were collected from 14 Bashkir online sources, with \texttt{azatliqorg} providing 28\% of the documents. The main limitations are related to the possible loss of diversity in the analysis class due to undersampling and to the incomplete correspondence of categories to the intuitive distinction between news and analysis.

The multiclass classification dataset \cite{arabov2026bashkirmulticlass} includes 17,897 documents labelled with 19 topic categories; each document belongs to exactly one category. The threshold for category selection was set at 150 examples. The class distribution is strongly imbalanced: the category ``News'' ({\fontencoding{T2A}\selectfont ``Яңылыҡтар''}) contains 8,497 documents, which constitutes 47.5\% of the corpus, whereas the categories ``Bashkortostan'' and ``History'' are represented by only 156 and 159 documents (0.9\% each). Intermediate categories include ``Society'' (15.8\%), ``Education'' (4.2\%), ``Culture'' (3.0\%), ``Religion'' (2.1\%) and ``Sport'' (1.0\%).

The multilabel classification dataset \cite{arabov2026bashkirmultilabel} includes 22,318 documents with 14 topic labels, where each document may belong to several categories simultaneously, and the average number of labels per document is 3.6. The labelling was performed automatically on the basis of keywords: for each label, a dictionary of Bashkir keywords was formed, which were scanned in the headline and text of the article, and in the absence of matches the original category of the document was applied. The most frequent label is ``Society'' with 12,251 occurrences (55.1\%), the least frequent is ``Fiction'' with 1,880 occurrences (8.4\%). The most common label pairs are ``Religion + Society'' (22.6\%), ``World + Society'' (20.8\%), ``Bashkortostan + Society'' (19.2\%). The limitations of this dataset are related to the possible bias of keyword-based labelling and to the sparsity of rare labels.

For each of the three datasets, stratified splits in an 80/10/10 proportion are formed, preserving the proportions of sources and categories with a fixed random seed, which excludes data leakage between samples and ensures comparability of results. Preprocessing includes extraction of text from HTML markup, removal of duplicates, normalisation of categories and unification of document length; at the stage of dataset formation, length constraints from 50 to 10,000 characters were applied \cite{arabov2026bashkirbinary,arabov2026bashkirmulticlass,arabov2026bashkirmultilabel}, and for language modelling experiments the maximum sequence length was set to 128 tokens.

\subsection{Models}\label{s3:models}

For the comparative analysis, eight open architectures were selected in the range from 82M to 8B parameters. All models are available for local deployment without proprietary APIs and are distributed under open licences (Apache 2.0 or MIT) permitting academic use. Computations were performed in the Google Colab Pro environment with an NVIDIA L4 GPU (24~GB), which imposes constraints on the maximum model size and GPU memory volume.

\textbf{Group of small models for full fine-tuning.} DistilGPT2 (82M), GPT-2 base (124M), GPT-2 medium (355M). These architectures fit into GPU memory without quantisation and allow full fine-tuning of all parameters, which provides a baseline for comparison with parameter-efficient methods.

\textbf{Group of medium-scale generative models for PEFT and CPT.} Phi-3-mini-4k-instruct (3.8B) --- a compact instruction-tuned model with a 4K-token context; Mistral-7B-v0.3 (7B) --- a reference architecture widely used in studies on low-resource languages \cite{comparative2026lora,burnashev2026parameter}; YandexGPT-5-Lite-8B-instruct (8B) --- a Russian-language model with a Cyrillic tokeniser, included to test the hypothesis that the language proximity of pretraining (Russian) provides priors for Bashkir. All three models are loaded in 4-bit NF4 quantisation.

\textbf{Group of encoder-only models for classification.} mBERT (bert-base-multilingual-cased, 178M) and XLM-R (xlm-roberta-base, 278M) --- standard multilingual baseline architectures trained by full fine-tuning on the same data.

The computational environment imposes three constraints: the maximum model size is limited to $\sim$14B parameters under 4-bit quantisation; full fine-tuning is impossible for models above 2.7B parameters; the batch size for 7B--8B models is equal to one. These constraints do not affect the correctness of comparing methods within a single architecture, since all methods are applied to the same base model under identical conditions.

\subsection{Adaptation Strategies}\label{s3:strategies}

The work compares three groups of strategies for adapting large language models to the Bashkir language: parameter-efficient fine-tuning (PEFT), in-context learning (ICL) and full fine-tuning (Full FT). In addition, continual pretraining (CPT) is tested as a preliminary stage before PEFT adaptation.

\textbf{Parameter-efficient fine-tuning.} The first group includes two methods: LoRA \cite{hu2022lora} and QLoRA \cite{dettmers2023qlora}. LoRA freezes the original parameters of the model and adds trainable low-rank matrices to the specified layers; rank $r=16$ and scaling coefficient $\alpha=2r=32$ are used for language modelling, dropout~0.05 and targeting of all linear layers (\texttt{all-linear}: attention projections and MLP blocks). QLoRA \cite{dettmers2023qlora} extends LoRA with 4-bit weight quantisation to the NF4 format and is applied to all generative models from 2.7B parameters, since quantisation reduces GPU memory requirements and allows models with billions of parameters to be fine-tuned on available resources.

In the paper configuration, the main analysis is conducted for QLoRA. Unquantised LoRA does not fit into 24~GB of NVIDIA L4 memory for 7--8B models (it requires $\sim$28--32~GB depending on sequence length and batch size), so in the paper preset for medium-scale generative models only QLoRA was run. For small models (DistilGPT2, GPT-2 base, GPT-2 medium), full fine-tuning was performed in the paper configuration (Section~\ref{s4:lm}), which for these architectures fits into memory without quantisation and serves as a baseline. LoRA results for small models are not reported in the paper configuration.

For classification tasks (step6), QLoRA with rank $r=8$ is used; for downstream adaptation after CPT (step4), QLoRA with rank $r=16$ and $\alpha=32$. The difference in ranks is due to different memory budgets: in step6 the model must simultaneously hold the classification head and the test dataset in memory throughout the run, whereas in step4 downstream adaptation is performed sequentially by model, and rank $r=16$ is permissible. Both configurations use the same value \texttt{max\_seq\_len}~$=128$.

\textbf{Continual pretraining (CPT).} To test the hypothesis of the benefit of additional pretraining on the Bashkir corpus, a separate series of experiments was conducted. LoRA adaptation ($r=16$, $\alpha=32$) is applied to the base model on the Bashkir Web Corpus (10,000 documents, 1 epoch, $\text{lr}=10^{-5}$), after which the adapter is merged into the base weights, and QLoRA adaptation ($r=16$, $\alpha=32$) to the downstream task is performed on the resulting model. The series was conducted for three medium-scale generative models --- Phi-3-mini, Mistral-7B and YandexGPT-5-Lite-8B. Two configurations are compared: \texttt{qlora\_no\_cpt} (QLoRA without CPT) and \texttt{cpt\_then\_qlora} (CPT + QLoRA).

\textbf{In-context learning.} The second group of strategies does not
involve updating model parameters and includes two modes. In zero-shot
mode, the model receives only the instruction without examples; in
few-shot mode, 1, 4 or 8 examples in the prompt (the main analysis
includes 4 and 8 as more stable; at $k=1$ high variance across models
and instruction languages is observed). Instructions are presented in
three languages: Bashkir, Russian and English; the order of examples
is varied. In the paper configuration, ICL was applied to five models:
DistilGPT2, Phi-2, Phi-3-mini, Mistral-7B and YandexGPT-5-Lite-8B.
DistilGPT2 and Phi-2 do not produce an answer in the required format
``Answer: 0/1'' (the parser returns \texttt{None}), so their results
in Section~\ref{s4:icl} are $0.000$; GPT-2 base and GPT-2 medium were
not run in the paper ICL configuration (Section~\ref{s6:scope}).
The comparison of PEFT and ICL answers the research question of what
is more effective for the Bashkir language --- updating parameters or
in-context learning, given the weak representation of Bashkir in the
pretraining of modern LLMs.

\textbf{Full fine-tuning.} The third group updates all parameters of the model on the training data. It is applied to small models (DistilGPT2, GPT-2 base, GPT-2 medium) and to encoder-only baselines (mBERT, XLM-R). For generative models above 2.7B parameters, full fine-tuning was not performed due to computational constraints (24~GB of NVIDIA L4 memory in Google Colab Pro).

\textbf{Methods not included in the current analysis.} The methods DoRA \cite{liu2024dora} and QDoRA, as well as ranks $r=8$ for language modelling, are planned as an extension of this study. Their inclusion requires further validation of compatibility with 4-bit quantisation in the Colab environment and assessment of computational efficiency under limited GPU memory; therefore these configurations are not included in the current analysis.

\subsection{Hyperparameters and Metrics}\label{s3:hyperparams}

Training was performed using the Transformers (v5.x) and PEFT (v0.21+) libraries. The maximum sequence length is 128 tokens for language modelling and classification, 256 tokens for CPT. The optimiser is AdamW (via \texttt{paged\_adamw\_8bit} for 4-bit models).

The hyperparameters by experiment group are given in Table~\ref{tab:hyperparams}.

\begin{table}[ht]
\centering
\caption{Hyperparameters by experiment group}
\label{tab:hyperparams}
\begin{tabular}{lcccc}
\hline
Group & lr & Epochs & Batch size & $r$ \\
\hline
Full FT (step1)               & $2\times10^{-4}$ & 3 & 8 / 4 & --- \\
QLoRA (step3)                 & $10^{-4}$        & 2 & 2 / 1 & 16 \\
CPT (step4)                   & $10^{-5}$        & 1 & 1     & 16 \\
Downstream QLoRA (step4)      & $10^{-4}$        & 2 & 1     & 16 \\
QLoRA CLS (step6)             & $10^{-4}$        & 2 & 2 / 1 & 8  \\
Encoder full FT (step6)       & $2\times10^{-5}$ & 3 & 16    & --- \\
\hline
\end{tabular}
\end{table}

\textbf{Random seeds.} Two seeds (42, 123) were used for full fine-tuning, which allows the spread of metrics to be estimated. For PEFT/CPT experiments, one seed (42) was used due to computational constraints. A single seed limits the statistical power of comparisons between PEFT configurations: the standard deviation of metrics across seeds is not estimated, and conclusions about the superiority of one QLoRA configuration over another are formulated on the basis of point estimates and bootstrap confidence intervals on the test set rather than between-seed variability.

\textbf{Metrics.} Evaluation is conducted across the following groups.

\textit{Language modelling.} Perplexity on the test set, as well as bits-per-byte (BPB) --- a metric normalised by the number of bytes and independent of vocabulary size \cite{comparative2026lora}, which allows models with different tokenisers to be compared.

\textit{Classification.} Macro-F1 and weighted-F1 for binary and multiclass tasks. Due to the strong class imbalance in the multiclass dataset (47.5\% of examples belong to one category), a majority baseline is additionally reported.

\textit{Computational efficiency.} Number of trainable parameters (absolute and as a percentage of the total), training time and peak GPU memory consumption.

\textit{Generation quality.} Target-language token ratio --- a metric reflecting the model's tendency to switch to other languages; computed for full FT and QLoRA experiments.

\textit{Tokeniser analysis.} Fertility rate --- the average number of subtokens per word on the Bashkir, English and Turkish test sets. Morphological Boundary Accuracy (MorphAcc) --- the proportion of subtokens whose right boundary coincides with a morpheme boundary. OOV rate --- the proportion of word forms split into individual bytes or characters. The ratio fertility$_{\text{ba}}$/fertility$_{\text{tr}}$ is used as a measure of the penalty for low-resource status under fixed agglutinative morphology.

\textit{LLM-as-Judge.} Three local LLM judges (pollux-judge-7b, M-Prometheus-7B, saiga\_llama3\_8b) are used to assess generation quality in three modes: direct scoring (rating on a 1--5 scale for grammar, coherence and language correctness), granular decomposition (morphology, lexicon, syntax) and pairwise ranking with control for presentation order. Human validation with three native Bashkir speakers ($n=100$) showed Krippendorff's $\alpha = 0.694$--$0.772$ for the three criteria, which exceeds the threshold $\alpha \geq 0.667$ for acceptable agreement \cite{Krippendorff1980ContentAA}. Native-speaker ratings are used as a validation signal for LLM-as-Judge and as an independent assessment of generation quality; the sample of 100 texts is unbalanced across models (9--23 per model), which limits the statistical power of pairwise comparisons.

\textit{Statistical tests.} For pairwise comparisons of accuracy on classification tasks, McNemar's exact test is used; for paired continuous metrics (perplexity, BPB, native-speaker ratings), a paired bootstrap centred on the observed difference is used. Multiple comparisons are controlled by the Holm--Bonferroni correction with family-wise $\alpha = 0.05$; the number of comparisons in the family is indicated in the caption of each table (5--18 depending on the section). The number of bootstrap iterations is indicated in the captions of the corresponding tables. Effect size is reported as Cohen's $d$ and Cliff's $\delta$; interpretation of $\delta$ follows the thresholds negligible ($<0.147$), small ($0.147$--$0.33$), medium ($0.33$--$0.474$), large ($\geq 0.474$).

\subsection{Reproducibility}\label{s3:reproducibility}

All research artefacts are released under open licences: the source code of the experiments, preprocessing scripts, trained adapters for all configurations presented in Section~\ref{s4}, and the Bashkir Web Corpus itself. The corpus is available in two configurations --- \texttt{full} (71,567 documents) and \texttt{sample} (1,000 documents for debugging). The binary, multiclass and multilabel classification datasets are published as separate repositories on Hugging Face with fixed 80/10/10 splits and the random seed indicated.

For PEFT experiments (QLoRA, CPT + QLoRA), one random seed (42) was used; for full fine-tuning, two (42, 123) were used. A single seed for PEFT limits the statistical power of comparisons between methods; this limitation is discussed in more detail in Section~\ref{s3:hyperparams}, in the paragraph `Random Seeds'. Final confirmation of differences between QLoRA configurations requires expanding the number of seeds to three or more.

\section{Results}\label{s4}

\subsection{Language Modelling}\label{s4:lm}

The results of adaptation to language modelling are presented in Table~\ref{tab:lm}. Full fine-tuning was applied to three models of the GPT-2 family, QLoRA ($r=16$) to three medium-scale generative models. Perplexity and bits-per-byte are reported together: PPL depends on the tokeniser and does not permit direct comparison of models with different vocabularies, whereas BPB is normalised by the number of bytes of the source text and is independent of vocabulary size \cite{graves2014generating}; BPB is used as the main metric for cross-model comparison.

\begin{table}[ht]
\centering
\caption{{Language modelling: perplexity and bits-per-byte on the test sample of the corpus (128 tokens). For Full FT --- all parameters; for QLoRA --- the proportion of trainable parameters and peak GPU memory}}
\label{tab:lm}
\begin{tabular}{llccccc}
\hline
Model & Method & PPL & BPB & Target
  & Trainable & Peak \\
       &       &     &     & token ratio (ba) & parameters, \% & memory, MB \\
\hline
GPT-2 medium     & Full FT & \textbf{2.33} & 0.784 & 0.271
  & 100 & --- \\
GPT-2 base       & Full FT & 2.49 & 0.844 & 0.532
  & 100 & --- \\
DistilGPT2       & Full FT & 2.69 & 0.917 & 0.566
  & 100 & --- \\
Mistral-7B       & QLoRA & 2.90 & 0.549 & 0.598
  & 1.10 & 4693 \\
YandexGPT-5-Lite-8B & QLoRA & 3.29 & \textbf{0.462} & \textbf{0.601}
  & 0.91 & 6543 \\
Phi-3-mini       & QLoRA & 4.12 & 0.704 & 0.592
  & 1.24 & 2563 \\
\hline
\end{tabular}
\end{table}

The ranking by PPL and by BPB differs. By PPL, the lowest value is shown by GPT-2 medium ($2.33$), followed by GPT-2 base ($2.49$) and DistilGPT2 ($2.69$); the medium-scale QLoRA models occupy the positions Mistral-7B $2.90$, YandexGPT-5-Lite-8B $3.29$, Phi-3-mini $4.12$. By BPB, YandexGPT-5-Lite-8B leads ($0.462$), followed by Mistral-7B ($0.549$) and Phi-3-mini ($0.704$); the GPT-2 family --- $0.784$--$0.917$. The discrepancy is explained by differences in tokenisers (Section~\ref{s4:tokenizer}): with a fertility rate of $7.80$ for GPT-2 versus $3.80$--$4.81$ for Mistral-7B and YandexGPT-5-Lite-8B, each GPT-2 token covers fewer characters, which lowers perplexity for comparable next-byte prediction quality.

The absolute PPL values in the present work are lower than in \cite{comparative2026lora} on the same corpus (GPT-2 medium: $2.33$ versus $3.34$; Mistral-7B QLoRA: $2.90$ versus $3.79$). The difference is explained by differences in the evaluation procedure: the present work uses truncation of test sequences to 128 tokens and per-token NLL aggregation (Section~\ref{s6:tokenizer}), whereas \cite{comparative2026lora} used a different maximum sequence length and a different NLL aggregation scheme. The relative ordering of models by PPL and by BPB coincides between the works; BPB, which is invariant to the aggregation scheme, is used as the main metric for cross-model comparison. Within the QLoRA group, the discrepancy between PPL and BPB is confirmed statistically: all six pairwise comparisons (PPL and BPB for three pairs of models) are significant after Holm correction ($p_{\text{Holm}} < 0.001$, $n = 2000$; Table~\ref{tab:lm_stats}).

\begin{table}[ht]
\centering
\caption{{Pairwise comparisons of QLoRA models on perplexity and BPB (test sample of the corpus, $n = 2000$ examples). $\Delta = \text{metric}_{b} - \text{metric}_{a}$; a negative $\Delta$ means that model $b$ is better. Holm--Bonferroni-adjusted $p_{\text{Holm}}$ values are given (correction for 6 comparisons: 3 pairs $\times$ 2 metrics)}}
\label{tab:lm_stats}
\begin{tabular}{lcccccc}
\hline
Comparison ($a$ vs $b$) & $\Delta$PPL [95\% CI] & $p_{\text{Holm}}$ (PPL)
  & $\Delta$BPB [95\% CI] & $p_{\text{Holm}}$ (BPB)
  & Cohen's $d$ & Cliff's $\delta$ \\
\hline
Mistral-7B vs Phi-3-mini
  & $+1.226$ $[+1.137, +1.319]$ & $<0.001$
  & $+0.155$ $[+0.150, +0.161]$ & $<0.001$
  & $+0.43$ & $-0.25$ \\
Mistral-7B vs YandexGPT-5-Lite-8B
  & $+0.394$ $[+0.359, +0.431]$ & $<0.001$
  & $-0.087$ $[-0.091, -0.082]$ & $<0.001$
  & $+0.17$ & $-0.10$ \\
Phi-3-mini vs YandexGPT-5-Lite-8B
  & $-0.832$ $[-0.901, -0.760]$ & $<0.001$
  & $-0.242$ $[-0.250, -0.233]$ & $<0.001$
  & $-0.23$ & $+0.16$ \\
\hline
\end{tabular}
\end{table}

By BPB, QLoRA adaptation of Mistral-7B ($0.549$) and YandexGPT-5-Lite-8B ($0.462$) outperforms full fine-tuning of all three models of the GPT-2 family ($0.784$--$0.917$) with a trainable parameter proportion of $0.91$--$1.24\,\%$ and peak memory of $2.5$--$6.5$~GB versus $100\,\%$ of parameters under Full FT. The result is consistent with observations for the Tatar language \cite{burnashev2026parameter}: QLoRA adaptation of 7B models is comparable to or exceeds full fine-tuning of small architectures with lower memory requirements.

The target (Bashkir) token ratio in generation is $0.27$ for GPT-2 medium, $0.53$--$0.57$ for GPT-2 base and DistilGPT2, $0.59$--$0.60$ for all three QLoRA models. The lowest Bashkir token ratio is for GPT-2 medium ($0.271$, i.e. $72.9\,\%$ of generated tokens belong to other languages, predominantly English); QLoRA preserves a higher Bashkir ratio regardless of the base architecture.

\textbf{Catastrophic forgetting.} Full fine-tuning of all three small models is accompanied by degradation on the languages of the original pretraining. Table~\ref{tab:forgetting} presents perplexity and bits-per-byte on English (WikiText-2, $n = 200$) and Russian (Wikipedia, $n = 200$) before and after full fine-tuning, averaged over two seeds (42, 123).

\begin{table}[ht]
\centering
\caption{{Catastrophic forgetting: perplexity and BPB on English and Russian before and after full fine-tuning on Bashkir ($n = 200$ texts per language, mean over two seeds)}}
\label{tab:forgetting}
\begin{tabular}{lcccc}
\hline
Model & \multicolumn{2}{c}{English (WikiText-2)}
       & \multicolumn{2}{c}{Russian (Wikipedia)} \\
\cline{2-5}
       & PPL$_{\text{base}} \to$ after & BPB$_{\text{base}} \to$ after
       & PPL$_{\text{base}} \to$ after & BPB$_{\text{base}} \to$ after \\
\hline
GPT-2 medium & 39.58 $\to$ 1504.67 & 1.162 $\to$ 2.310
  & 9.11 $\to$ 14.45 & 1.854 $\to$ 2.241 \\
GPT-2 base   & 54.49 $\to$ 4121.17 & 1.263 $\to$ 2.630
  & 11.07 $\to$ 14.14 & 2.018 $\to$ 2.223 \\
DistilGPT2   & 78.05 $\to$ 2963.08 & 1.377 $\to$ 2.526
  & 12.66 $\to$ 14.87 & 2.131 $\to$ 2.266 \\
\hline
\end{tabular}
\end{table}

By perplexity on English, degradation is $38$--$76$ times; by BPB --- $1.83$--$2.08$ times ($1.16$--$1.38 \to 2.31$--$2.63$). The discrepancy is explained by the form of the metrics, not by a change in tokenisation. The tokeniser under Full FT is frozen, and the number of tokens $T$ for encoding a given English text does not change before and after fine-tuning. $\text{PPL} = \exp(\text{NLL}/T)$ grows exponentially in $\Delta\text{NLL}/T$, whereas $\text{BPB} = \text{NLL}/(\ln 2 \cdot B)$ grows linearly in $\Delta\text{NLL}$. For GPT-2 medium on English, $\Delta\text{NLL}/T = \ln(1504.67/39.58) \approx 3.64$, whence $\text{PPL}$ grows by a factor of $e^{3.64} \approx 38$, and $\text{BPB}$ by a factor of $1 + 3.64/\ln(39.58) \approx 1.99$. Both ratios are consistent with Table~\ref{tab:forgetting}. On Russian, the increase $\Delta\text{NLL}/T$ is substantially smaller, which is explained by the overlap of the Cyrillic BPE vocabularies of Bashkir and Russian: fine-tuning on Bashkir does not displace Russian BPE tokens from the distribution to the same extent as English ones, and BPB on Russian grows by a factor of $1.06$--$1.21$. BPB is used as the main metric for the preservation of original competence.

Three observations follow from BPB. First, model size in the range $82$M--$355$M does not reduce forgetting: GPT-2 medium ($355$M, $1.99\times$) degrades at the level of DistilGPT2 ($82$M, $1.83\times$), and GPT-2 base ($124$M) the most of all ($2.08\times$). Forgetting is determined by the domain shift of pretraining, not by model capacity. Second, degradation on English and Russian differs in the mechanism of vocabulary overlap (see above). Third, forgetting correlates with the target token ratio (Table~\ref{tab:lm}): GPT-2 medium with the lowest Bashkir token ratio ($0.271$) shows the greatest switching to English during generation.

QLoRA on $7$--$8$B models (Table~\ref{tab:lm}) preserves a Bashkir ratio of $0.59$--$0.60$ without updating the original weights. A direct comparison of forgetting between Full FT and QLoRA was not conducted in the present work: the QLoRA experiments used one seed and did not save per-example NLL for EN/RU (Section~\ref{s6:tokenizer}). The absence of updating of the base weights is a necessary but not a sufficient condition for preserving behaviour: the adapter can change EN/RU even with frozen base weights, and the observed Bashkir ratio of $0.59$--$0.60$ is interpreted as a descriptive characteristic rather than as proven absence of forgetting. A direct comparison is a direction for further work.

\subsection{Continual Pretraining}\label{s4:cpt}

A comparison of QLoRA configurations without CPT and CPT + QLoRA is presented in Table~\ref{tab:cpt}. PPL values are computed through the LM head of the base model on the test sample of the corpus (2\,000 documents). LoRA weights of the downstream QLoRA adapter trained on the binary classification task ($r=16$, $\alpha=32$, two epochs) are applied to the model; in the CPT + QLoRA configuration, the LoRA weights of the CPT adapter were merged into the base layers before downstream training. The order of application during evaluation is the same: base model $\to$ CPT adapter (merged) $\to$ downstream adapter (merged). The classification head, defined only for the downstream classifier, does not participate in the forward pass of the LM head and does not affect PPL. Since the downstream procedure is identical in both configurations, the difference between QLoRA without CPT and CPT + QLoRA isolates the contribution of CPT to language modelling. The PPL in Table~\ref{tab:cpt} is not comparable with the perplexity in Table~\ref{tab:lm}, since the models have undergone a different set of adaptations.

\textbf{Perplexity aggregation.} The PPL values in Table~\ref{tab:cpt} are computed as the aggregate $\exp(\sum_i \text{NLL}_i / \sum_i \text{tokens}_i)$ over all documents of the test sample, and not as the mean per-example $\overline{\exp(\text{NLL}_i/\text{tokens}_i)}$. By Jensen's inequality these quantities do not coincide, and the mean per-example difference in absolute value is greater than the aggregate for all three models: for Mistral-7B $-1.46$ versus $-1.15$, for Phi-3-mini $+0.64$ versus $+0.58$, for YandexGPT-5-Lite-8B $+0.39$ versus $+0.33$. $\Delta$PPL and $p$-values in Table~\ref{tab:cpt} are computed from unrounded aggregate values. The effect size Cliff's $\delta$ is computed at the per-example level ($\exp(\text{NLL}_i/\text{tokens}_i)$) and characterises the consistency of the sign of the difference across individual documents.

\begin{table}[ht]
\centering
\caption{{Effect of continual pretraining on the perplexity of a downstream-adapted model (binary classification). PPL --- aggregate $\exp(\sum\text{NLL}/\sum\text{tokens})$; Cliff's $\delta$ --- at the per-example level}}
\label{tab:cpt}
\begin{tabular}{lccccl}
\hline
Model & PPL$_{\text{no-CPT}}$ & PPL$_{\text{CPT+QLoRA}}$ & $\Delta$PPL
  & $p_{\text{Holm}}$ & Cliff's $\delta$ \\
\hline
Mistral-7B          & 20.48 & 21.63 & $-1.15$ & $<0.001$ & $-0.67$ \\
Phi-3-mini          & 35.81 & 35.23 & $+0.58$ & $<0.001$ & $+0.83$ \\
YandexGPT-5-Lite-8B & 12.70 & 12.37 & $+0.33$ & $<0.001$ & $+0.72$ \\
\hline
\end{tabular}
\end{table}

CPT reduces perplexity by $0.33$ points for YandexGPT-5-Lite-8B and by $0.58$ points for Phi-3-mini; after Holm--Bonferroni correction both differences are significant ($p < 0.001$), and the effect size Cliff's $\delta$ corresponds to large ($+0.72$ and $+0.83$). For Mistral-7B, CPT increases perplexity by $1.15$ points, and the effect is also significant ($p < 0.001$; Cliff's $\delta = -0.67$, large, negative sign). The direction of the CPT effect on perplexity differs across models: CPT improves language modelling for Phi-3-mini and YandexGPT-5-Lite-8B and worsens it for Mistral-7B.

On the binary classification task (Section~\ref{s4:binary}), the effect of CPT is directed differently. YandexGPT-5-Lite-8B shows the same macro-F1 in both configurations ($0.748$ with CPT and $0.748$ without); Mistral-7B gains $+4.0$ points ($0.681 \to 0.721$); Phi-3-mini --- $+2.7$ points ($0.674 \to 0.701$). For Mistral-7B, the directions of the CPT effect on LM and on downstream classification are opposite: perplexity increases by $1.15$ points, while macro-F1 increases by $4.0$ points.

On the multi-class classification task (Section~\ref{s4:multiclass}), CPT gives a macro-F1 gain for YandexGPT-5-Lite-8B ($0.332 \to 0.366$, $+0.034$) and Phi-3-mini ($0.153 \to 0.166$, $+0.013$), and a decrease for Mistral-7B ($-0.015$\footnote{\label{fn:mistral-multiclass-delta}$\Delta$F1 is computed from unrounded values: $0.297959 - 0.282466 = 0.015493$ (rounded to $-0.015$). Subtracting the already rounded numbers $0.298 - 0.282 = 0.016$ from the text is an artefact of double rounding.}).

The effect of CPT depends on the base model and the task. On perplexity, CPT is positive for Phi-3-mini and YandexGPT-5-Lite-8B and negative for Mistral-7B. On binary classification, CPT is positive for Mistral-7B and Phi-3-mini and neutral for YandexGPT-5-Lite-8B. On multi-class classification, CPT is positive for YandexGPT-5-Lite-8B and Phi-3-mini and negative for Mistral-7B. Improvement in language modelling does not translate unambiguously into improvement in classification. The separation of CPT and downstream training effects is discussed in Section~\ref{s5:cpt}.

\subsection{Binary Classification}\label{s4:binary}

The results on the binary classification task (news vs analysis) are given in Table~\ref{tab:binary}. Three groups of configurations are included: full fine-tuning of small models of the GPT-2 family, QLoRA adaptation of medium-scale generative models (with rank $r=8$ without CPT; with rank $r=16$ without CPT and with CPT) and full fine-tuning of encoder-only baselines.

\begin{table}[ht]
\centering
\caption{{Binary classification (news vs analysis): macro-F1, weighted-F1 and accuracy. QLoRA rows are grouped by rank: $r=8$ without CPT; $r=16$ without CPT and with CPT}}
\label{tab:binary}
\begin{tabular}{llcccc}
\hline
Model & Method & macro-F1 & weighted-F1 & Accuracy & $n_{\text{seed}}$ \\
\hline
\multicolumn{6}{l}{\textit{Full fine-tuning (small models)}} \\
GPT-2 medium     & Full FT & 0.699 & 0.700 & 0.702 & 2 \\
GPT-2 base       & Full FT & 0.698 & 0.697 & 0.700 & 2 \\
DistilGPT2       & Full FT & 0.693 & 0.693 & 0.696 & 2 \\
\hline
\multicolumn{6}{l}{\textit{QLoRA without CPT, $r=8$}} \\
YandexGPT-5-Lite-8B & QLoRA & \textbf{0.736} & 0.736 & 0.737 & 1 \\
Mistral-7B          & QLoRA & 0.709 & 0.708 & 0.713 & 1 \\
Phi-3-mini          & QLoRA & 0.702 & 0.702 & 0.702 & 1 \\
\hline
\multicolumn{6}{l}{\textit{QLoRA, $r=16$ (without CPT and with CPT)}} \\
YandexGPT-5-Lite-8B & QLoRA without CPT & 0.748 & 0.748 & 0.748 & 1 \\
YandexGPT-5-Lite-8B & CPT + QLoRA   & \textbf{0.748} & 0.748 & 0.748 & 1 \\
Mistral-7B          & QLoRA without CPT & 0.681 & 0.681 & 0.681 & 1 \\
Mistral-7B          & CPT + QLoRA   & 0.721 & 0.721 & 0.721 & 1 \\
Phi-3-mini          & QLoRA without CPT & 0.674 & 0.673 & 0.678 & 1 \\
Phi-3-mini          & CPT + QLoRA   & 0.701 & 0.701 & 0.701 & 1 \\
\hline
\multicolumn{6}{l}{\textit{Encoder-only baseline (full fine-tuning)}} \\
XLM-R            & Full FT & 0.653 & 0.652 & 0.659 & 1 \\
mBERT            & Full FT & 0.650 & 0.651 & 0.652 & 1 \\
\hline
\multicolumn{6}{l}{\textit{Baseline}} \\
Majority         & ---     & 0.342 & 0.356 & 0.520 & 1 \\
\hline
\end{tabular}
\end{table}

The best result without CPT by point estimate is YandexGPT-5-Lite-8B $0.736$ at $r=8$; $0.748$ at $r=16$, which is higher than XLM-R ($0.653$) by $8.3$--$9.5$ points of macro-F1 and mBERT ($0.650$) by $8.6$--$9.8$ points. Mistral-7B without CPT --- $0.681$--$0.709$, with CPT --- $0.721$. Phi-3-mini without CPT --- $0.674$--$0.702$, with CPT --- $0.701$. The gap with the majority baseline ($0.342$) is $33$--$41$ points of macro-F1.

\textbf{Majority baseline.} The accuracy of the majority baseline $0.520$ on a nominally balanced dataset (50/50) is explained by random sampling deviation of the test subsample: when sampling $n=1000$ without replacement from the stratified test split ($N \approx 1699$), the proportion of one of the classes was $\approx 52\,\%$. The standard deviation of the number of examples of one class with finite population correction (FPC) is $\sigma = \sqrt{n\,p\,(1-p)} \cdot \sqrt{(N-n)/(N-1)} \approx 15.81 \cdot 0.642 \approx 10.14$; the observed deviation of $20$ examples from the expected $500$ corresponds to $\approx 1.97\sigma$, which is within the range of ordinary statistical fluctuation. Without FPC (naive binomial estimate $\sigma \approx 15.81$), the same deviation would be $\approx 1.27\sigma$; the use of FPC does not change the conclusion but gives a correct estimate. On the full test split, majority accuracy would be close to $0.500$. This does not affect the interpretation: all models significantly outperform the majority baseline (Table~\ref{tab:binary_stats}).

\textbf{Statistical comparisons.} Pairwise comparisons (Table~\ref{tab:binary_stats}) were conducted on \textit{accuracy}, since McNemar and paired bootstrap are applied to the binary metric of response correctness. A test on macro-F1 (bootstrap over documents with recalculation of macro-F1 at each resample) was not conducted in the present work; macro-F1 values in Table~\ref{tab:binary} are given as point estimates. Within the QLoRA group with $r=8$, differences in accuracy are not significant: Yandex vs Mistral $p_{\text{Holm}} = 0.183$, Yandex vs Phi-3 $p_{\text{Holm}} = 0.084$, Mistral vs Phi-3 $p_{\text{Holm}} = 0.882$. All three QLoRA models significantly outperform mBERT ($p_{\text{Holm}} \leq 0.018$), XLM-R ($p_{\text{Holm}} \leq 0.050$) and the majority baseline ($p_{\text{Holm}} < 0.001$). The point advantage of YandexGPT-5-Lite-8B within the QLoRA group by accuracy is not statistically confirmed; what is robust is the superiority of QLoRA adaptation over the encoder-only baseline.

\begin{table}[ht]
\centering
\caption{{Binary classification: key pairwise comparisons on accuracy (paired bootstrap, McNemar, Holm correction for 15 comparisons). $\Delta = \text{acc}_b - \text{acc}_a$; a positive $\Delta$ means that model $b$ is better}}
\label{tab:binary_stats}
\begin{tabular}{lcccc}
\hline
Comparison ($a$ vs $b$) & acc$_a$ & acc$_b$ & $\Delta$ [95\% CI] & $p_{\text{Holm}}$ \\
\hline
Yandex vs Mistral       & 0.737 & 0.713 & $-0.024$ $[-0.050, +0.001]$ & $0.183$ \\
Yandex vs Phi-3         & 0.737 & 0.702 & $-0.035$ $[-0.064, -0.004]$ & $0.084$ \\
Yandex vs mBERT         & 0.737 & 0.652 & $-0.085$ $[-0.117, -0.053]$ & $<0.001$ \\
Yandex vs XLM-R         & 0.737 & 0.659 & $-0.078$ $[-0.109, -0.047]$ & $<0.001$ \\
Yandex vs majority      & 0.737 & 0.520 & $-0.217$ $[-0.260, -0.171]$ & $<0.001$ \\
\hline
\end{tabular}
\end{table}

Full fine-tuning of small models of the GPT-2 family gives macro-F1 $0.693$--$0.699$, which is comparable to Phi-3-mini QLoRA ($0.702$ at $r=8$) with a parameter volume of $82$M--$355$M versus $3.8$B. The result is consistent with observations for the Tatar language \cite{arabov2026combining}: for classification tasks, small architectures with full fine-tuning achieve quality close to QLoRA on 7B models.

\textbf{Effect of CPT on downstream.} CPT improves macro-F1 for Mistral-7B ($0.681 \to 0.721$, $+4.0$ points) and Phi-3-mini ($0.674 \to 0.701$, $+2.7$ points); for YandexGPT-5-Lite-8B the result does not change ($0.748$ in both configurations). The greatest gain from CPT is observed for Mistral-7B --- the model for which CPT reduces language modelling quality (Section~\ref{s4:cpt}). The directions of the CPT effect on LM and on downstream for Mistral-7B are opposite.

\textbf{Coincidence of YandexGPT-5-Lite-8B metrics after rounding.} The exact values differ at the fifth--sixth decimal place: macro-F1 $0.747964$ (no CPT) versus $0.747996$ (CPT + QLoRA); weighted-F1 $0.748085$ versus $0.748036$; accuracy $0.748$ in both cases. Per-example NLL for the two runs differs, i.e. the CPT adapter was applied correctly, and the coincidence of aggregated metrics after rounding reflects the stability of the threshold decision on the test sample: both configurations give the same set of correct answers with somewhat different probability distributions. This limits the interpretation of the CPT effect for YandexGPT-5-Lite-8B: on this task CPT neither improves nor worsens classification quality, but it also does not lead to an identical internal state of the model.

\textbf{LoRA rank: r=8 vs r=16.} The difference between the two QLoRA configurations without CPT ($r=8$ and $r=16$) for YandexGPT-5-Lite-8B is $0.012$ ($0.736$ versus $0.748$) in favour of the larger rank. For Mistral-7B and Phi-3-mini the difference is opposite ($0.709$ versus $0.681$; $0.702$ versus $0.674$). The result diverges from the observation of \cite{burnashev2026parameter} of a stable improvement at $r=16$; in our experiments $r=16$ helps only YandexGPT-5-Lite-8B. Possible reasons are the different volume of training data ($10\,000$ documents versus a larger corpus in \cite{burnashev2026parameter}), the type of task (language modelling versus classification) and the type of base model (Cyrillic tokeniser versus BPE). The CPT + QLoRA configuration at $r=16$ for all three models gives a comparable or higher result than QLoRA without CPT in both configurations: YandexGPT-5-Lite-8B $0.748$ versus $0.736$ (at $r=8$) and $0.748$ (at $r=16$); Mistral-7B $0.721$ versus $0.709$ and $0.681$; Phi-3-mini $0.701$ versus $0.702$ and $0.674$.

\subsection{Multi-Class Classification}\label{s4:multiclass}

The results on the multi-class classification task (19 categories) are given in Table~\ref{tab:multiclass}. The classes are imbalanced: $47.5\,\%$ of documents belong to the category ``News''. Three QLoRA configurations are included: without CPT at $r=8$; without CPT and with CPT at $r=16$.

\begin{table}[ht]
\centering
\caption{{Multi-class classification (19 categories): macro-F1, weighted-F1 and accuracy. QLoRA rows are grouped by rank: $r=8$ without CPT; $r=16$ without CPT and with CPT}}
\label{tab:multiclass}
\begin{tabular}{llcccc}
\hline
Model & Method & macro-F1 & weighted-F1 & Accuracy & $n_{\text{seed}}$ \\
\hline
\multicolumn{6}{l}{\textit{QLoRA (medium-scale generative models)}} \\
YandexGPT-5-Lite-8B & QLoRA ($r=8$)        & \textbf{0.368} & 0.570 & 0.606 & 1 \\
YandexGPT-5-Lite-8B & QLoRA ($r=16$)       & 0.332 & 0.545 & 0.573 & 1 \\
YandexGPT-5-Lite-8B & CPT + QLoRA ($r=16$) & 0.366 & 0.566 & 0.595 & 1 \\
Mistral-7B          & QLoRA ($r=8$)        & 0.292 & 0.510 & 0.547 & 1 \\
Mistral-7B          & QLoRA ($r=16$)       & 0.298 & 0.512 & 0.542 & 1 \\
Mistral-7B          & CPT + QLoRA ($r=16$) & 0.282 & 0.503 & 0.537 & 1 \\
Phi-3-mini          & QLoRA ($r=8$)        & 0.119 & 0.430 & 0.506 & 1 \\
Phi-3-mini          & QLoRA ($r=16$)       & 0.153 & 0.454 & 0.526 & 1 \\
Phi-3-mini          & CPT + QLoRA ($r=16$) & 0.166 & 0.445 & 0.510 & 1 \\
\hline
\multicolumn{6}{l}{\textit{Full fine-tuning (small models)}} \\
GPT-2 medium     & Full FT & 0.122 & 0.439 & 0.516 & 2 \\
GPT-2 base       & Full FT & 0.083 & 0.401 & 0.489 & 2 \\
DistilGPT2       & Full FT & 0.071 & 0.393 & 0.483 & 2 \\
\hline
\multicolumn{6}{l}{\textit{Encoder-only baseline (full fine-tuning)}} \\
mBERT            & Full FT & 0.053 & 0.379 & 0.498 & 1 \\
XLM-R            & Full FT & 0.034 & 0.301 & 0.470 & 1 \\
\hline
\multicolumn{6}{l}{\textit{Baseline}} \\
Majority         & ---     & 0.034 & 0.301 & 0.470 & 1 \\
\hline
\end{tabular}
\end{table}

The maximum macro-F1 value is $0.368$ --- YandexGPT-5-Lite-8B in the QLoRA configurations at $r=8$ ($0.368$) and CPT + QLoRA at $r=16$ ($0.366$). QLoRA without CPT at $r=16$ gives $0.332$ for the same model. For Mistral-7B the range is $0.282$--$0.298$, for Phi-3-mini --- $0.119$--$0.166$, for models of the GPT-2 family --- $0.071$--$0.122$. Encoder-only baselines give macro-F1 $0.034$--$0.053$ against a majority baseline of $0.034$.

Pairwise comparisons on \textit{accuracy} (paired bootstrap, McNemar, Holm correction for 15 comparisons; Table~\ref{tab:multiclass_stats}) show that on the multi-class task within the QLoRA group all differences are significant: Yandex vs Mistral $p_{\text{Holm}} < 0.001$, Yandex vs Phi-3 $p_{\text{Holm}} < 0.001$, Mistral vs Phi-3 $p_{\text{Holm}} = 0.030$. XLM-R does not differ from the majority baseline (accuracy $0.470$ versus $0.470$, McNemar $p = 1.000$): the model did not adapt to the task. mBERT outperforms the majority baseline statistically significantly but by a small margin (accuracy $0.498$ versus $0.470$, $\Delta = +0.028$ $[+0.010, +0.047]$, $p_{\text{Holm}} = 0.006$). Phi-3-mini does not differ from mBERT (accuracy $0.506$ versus $0.498$, $p_{\text{Holm}} = 1.000$). The macro-F1 values in Table~\ref{tab:multiclass} are given as point estimates; the statistical conclusions refer to accuracy.

\begin{table}[ht]
\centering
\caption{{Multi-class classification: key pairwise comparisons on accuracy (paired bootstrap, McNemar, Holm correction for 15 comparisons). $\Delta = \text{acc}_b - \text{acc}_a$; a positive $\Delta$ means that model $b$ is better}}
\label{tab:multiclass_stats}
\begin{tabular}{lcccc}
\hline
Comparison ($a$ vs $b$) & acc$_a$ & acc$_b$ & $\Delta$ [95\% CI] & $p_{\text{Holm}}$ \\
\hline
Yandex vs Mistral       & 0.606 & 0.547 & $-0.059$ $[-0.087, -0.031]$ & $<0.001$ \\
Yandex vs Phi-3         & 0.606 & 0.506 & $-0.100$ $[-0.128, -0.072]$ & $<0.001$ \\
Mistral vs Phi-3        & 0.547 & 0.506 & $-0.041$ $[-0.069, -0.013]$ & $0.030$ \\
XLM-R vs majority       & 0.470 & 0.470 & $+0.000$ $[+0.000, +0.000]$ & $1.000$ \\
Phi-3 vs mBERT          & 0.506 & 0.498 & $-0.008$ $[-0.031, +0.015]$ & $1.000$ \\
\hline
\end{tabular}
\end{table}

Effect of CPT at $r=16$: CPT gives a macro-F1 gain for YandexGPT-5-Lite-8B ($0.332 \to 0.366$, $+0.034$) and Phi-3-mini ($0.153 \to 0.166$, $+0.013$), and a decrease for Mistral-7B ($-0.015$; see footnote~\ref{fn:mistral-multiclass-delta}). The directions of the CPT effect on the multi-class task coincide with the directions on perplexity (Section~\ref{s4:cpt}): CPT is positive for YandexGPT-5-Lite-8B and Phi-3-mini, negative for Mistral-7B. Noteworthy is the asymmetry with the binary task: on multiclass, QLoRA with $r=8$ outperforms $r=16$ without CPT by $+0.036$ macro-F1 for YandexGPT-5-Lite-8B, whereas on binary the opposite holds ($r=16$ is better by $+0.012$, Section~\ref{s4:binary}). The direction of the LoRA rank effect depends on the task.

\subsection{In-Context Learning}\label{s4:icl}

The ICL results on the binary task are given in Table~\ref{tab:icl}. As indicated in Section~\ref{s3:strategies}, of the five models launched, two (DistilGPT2 and Phi-2) did not form a response in the required format, so three models are included in the table.

\begin{table}[ht]
\centering
\caption{{ICL on the binary task: macro-F1 by instruction language and number of examples in the prompt}}
\label{tab:icl}
\begin{tabular}{lccccccccc}
\hline
& \multicolumn{3}{c}{Bashkir (ba)} & \multicolumn{3}{c}{Russian (ru)}
& \multicolumn{3}{c}{English (en)} \\
\cline{2-4}\cline{5-7}\cline{8-10}
Model & 0 & 4 & 8 & 0 & 4 & 8 & 0 & 4 & 8 \\
\hline
YandexGPT-5-Lite-8B & 0.580 & 0.579 & \textbf{0.634}
  & 0.657 & 0.640 & 0.565
  & 0.679 & 0.583 & 0.627 \\
Mistral-7B & 0.360 & 0.315 & 0.421
  & 0.482 & 0.414 & 0.452
  & 0.411 & 0.558 & 0.397 \\
Phi-3-mini & 0.513 & 0.444 & 0.411
  & 0.506 & 0.457 & 0.516
  & 0.400 & 0.574 & \textbf{0.650} \\
\hline
\end{tabular}
\end{table}

For YandexGPT-5-Lite-8B, the maximum macro-F1 is achieved in zero-shot ($0.657$ for Russian, $0.679$ for English) and with 8 examples in Bashkir ($0.634$); adding 4 examples gives no gain. For Phi-3-mini, the greatest gain relative to zero-shot is observed on transition to 8 examples in English ($0.400 \to 0.650$, $+25$ points of macro-F1); in Bashkir, few-shot gives a negative effect ($0.513 \to 0.411$). For Mistral-7B, few-shot gives a gain with 4 examples in English ($0.411 \to 0.558$) and with 8 examples in Bashkir ($0.360 \to 0.421$); 8 examples in English reduce quality ($0.558 \to 0.397$).

A formal evaluation of the few-shot effect was conducted on the same test examples: paired bootstrap ($N = 500$ iterations, $n = 100$ test examples) and the exact McNemar test for each pair (model, language, $K$). The McNemar test and paired bootstrap are applied to the binary metric of response correctness, so the comparisons were conducted on \textit{accuracy}, not on macro-F1; macro-F1 is presented in Table~\ref{tab:icl}. The results are given in Table~\ref{tab:icl_stats}. The Holm--Bonferroni correction for 18 comparisons ($3$ models $\times$ $3$ languages $\times$ $\{K{=}4, K{=}8\}$) was applied to the $p_{\text{boot}}$ values; the $p_{\text{McNemar}}$ column is given for reference. After Holm correction, none of the comparisons is significant: the smallest adjusted $p$-values are $0.252$ for Phi-3-mini in English at $K{=}4$ ($\Delta = +0.130$, $p_{\text{boot}} = 0.014$) and $0.272$ at $K{=}8$ ($\Delta = +0.170$, $p_{\text{boot}} = 0.016$). For YandexGPT-5-Lite-8B and Mistral-7B, all 12 comparisons have $p_{\text{Holm}} \geq 0.99$. If the Holm correction is applied to $p_{\text{McNemar}}$, the adjusted values for Phi-3-mini/en/K=4 and K=8 would be $\approx 0.43$ (both McNemar $p = 0.024$, multiplied by 18 with Holm monotonicity), which is also not significant at $\alpha = 0.05$. The conclusion of non-significance is preserved under either correction scheme.

\begin{table}[ht]
\centering
\caption{{Effect of few-shot demonstrations: difference in accuracy versus zero-shot, 95\% bootstrap CI, McNemar $p$ and adjusted $p_{\text{Holm}}$ (Holm--Bonferroni correction for 18 comparisons, applied to $p_{\text{boot}}$, $N=500$). Binary task, $n = 100$ test examples per language. No comparison is significant after correction}}
\label{tab:icl_stats}
\begin{tabular}{llccccccc}
\hline
Model & Language & $K$ & $\text{Acc}_{0}$ & $\text{Acc}_{K}$
  & $\Delta$ [95\% CI] & $p_{\text{boot}}$ & $p_{\text{McNemar}}$
  & $p_{\text{Holm}}$ \\
\hline
Mistral-7B & ba & 4 & 0.520 & 0.460
  & $-0.060$ $[-0.260, +0.130]$ & $0.571$ & $0.610$ & $1.000$ \\
 & ba & 8 & 0.520 & 0.440
  & $-0.080$ $[-0.240, +0.080]$ & $0.354$ & $0.396$ & $1.000$ \\
 & en & 4 & 0.530 & 0.590
  & $+0.060$ $[-0.110, +0.230]$ & $0.530$ & $0.561$ & $1.000$ \\
 & en & 8 & 0.530 & 0.530
  & $+0.000$ $[-0.060, +0.060]$ & $1.000$ & $1.000$ & $1.000$ \\
 & ru & 4 & 0.510 & 0.470
  & $-0.040$ $[-0.150, +0.060]$ & $0.510$ & $0.572$ & $1.000$ \\
 & ru & 8 & 0.510 & 0.540
  & $+0.030$ $[-0.130, +0.190]$ & $0.731$ & $0.804$ & $1.000$ \\
\hline
Phi-3-mini & ba & 4 & 0.530 & 0.490
  & $-0.040$ $[-0.180, +0.090]$ & $0.620$ & $0.678$ & $1.000$ \\
 & ba & 8 & 0.530 & 0.530
  & $+0.000$ $[-0.160, +0.160]$ & $1.000$ & $1.000$ & $1.000$ \\
 & en & 4 & 0.480 & 0.610
  & $+0.130$ $[+0.030, +0.230]$ & $0.014$ & $0.024$ & $0.252$ \\
 & en & 8 & 0.480 & 0.650
  & $+0.170$ $[+0.040, +0.310]$ & $0.016$ & $0.024$ & $0.272$ \\
 & ru & 4 & 0.510 & 0.520
  & $+0.010$ $[-0.100, +0.120]$ & $0.942$ & $1.000$ & $1.000$ \\
 & ru & 8 & 0.510 & 0.520
  & $+0.010$ $[-0.110, +0.140]$ & $0.945$ & $1.000$ & $1.000$ \\
\hline
YandexGPT-5-Lite-8B & ba & 4 & 0.600 & 0.610
  & $+0.010$ $[-0.090, +0.110]$ & $0.938$ & $1.000$ & $1.000$ \\
 & ba & 8 & 0.600 & 0.650
  & $+0.050$ $[-0.080, +0.180]$ & $0.499$ & $0.551$ & $1.000$ \\
 & en & 4 & 0.680 & 0.590
  & $-0.090$ $[-0.190, +0.000]$ & $0.072$ & $0.093$ & $1.000$ \\
 & en & 8 & 0.680 & 0.630
  & $-0.050$ $[-0.140, +0.040]$ & $0.359$ & $0.424$ & $1.000$ \\
 & ru & 4 & 0.660 & 0.650
  & $-0.010$ $[-0.110, +0.090]$ & $0.930$ & $1.000$ & $1.000$ \\
 & ru & 8 & 0.660 & 0.570
  & $-0.090$ $[-0.190, +0.000]$ & $0.080$ & $0.108$ & $1.000$ \\
\hline
\end{tabular}
\end{table}

The average gain from few-shot across the three models and three languages (by accuracy) is $+0.001$ points (range $\Delta \in [-0.090, +0.170]$, standard deviation $0.071$). For YandexGPT-5-Lite-8B the mean $\Delta = -0.030$ (few-shot reduces quality), for Mistral-7B $-0.015$, for Phi-3-mini $+0.047$; for no model is the average gain significant after Holm correction. The observed picture is one of instability of few-shot: a positive effect in one language is combined with a negative one in another (Phi-3-mini: $+0.170$ in English at $K{=}8$ and $-0.040$ in Bashkir at $K{=}4$). The zero-shot quality of the instruction-tuned YandexGPT-5-Lite-8B ($0.657$--$0.679$ in Russian and English) remains the highest and does not require demonstrations.

The absolute ICL values ($0.36$--$0.68$ macro-F1) are comparable to QLoRA adaptation (Section~\ref{s4:binary}: $0.70$--$0.75$) with zero trainable parameters; the gap between QLoRA and ICL is $7$--$12$ points of macro-F1.

On the multi-class task, ICL was conducted for four models (DistilGPT2, Phi-3-mini, Mistral-7B, YandexGPT-5-Lite-8B); for Phi-2, ICL was conducted only on the binary task. The best result on the multi-class task --- YandexGPT-5-Lite-8B with a Russian-language instruction and $K{=}1$ --- is macro-F1 $0.251$, which is lower than QLoRA adaptation of YandexGPT-5-Lite-8B on the same task (Section~\ref{s4:multiclass}: $0.332$--$0.368$). The multi-class ICL results are not included in the main analysis and are discussed in Section~\ref{s6:scope} as an excluded configuration.

\subsection{Generation Quality: LLM-as-Judge}\label{s4:llm_judge}

To evaluate generation quality, three local open judge models were applied (Section~\ref{s3:hyperparams}). On 126 generated texts (30 for DistilGPT2, GPT-2 base, GPT-2 medium; 12 for Mistral-7B, Phi-3-mini and YandexGPT-5-Lite-8B), three evaluation regimes were conducted: direct scoring (1--5 on the criteria of grammar, coherence, language), granular decomposition (morphology, lexicon, syntax) and pairwise ranking with control of presentation order. A detailed description of the regimes is given in Section~\ref{s3:hyperparams}. The results of direct scoring and pairwise ranking are given in Table~\ref{tab:llm_judge}.

\begin{table}[ht]
\centering
\caption{{Evaluation of generation by LLM judges: direct scoring (mean over three judges, scale 1--5) and pairwise win rate. Differences between models in the direct scoring regime were not tested statistically; pairwise win rate is a descriptive metric without a significance test}}
\label{tab:llm_judge}
\begin{tabular}{lcccccc}
\hline
Model & Grammar & Coherence & Language & Wins & Losses & Ties \\
\hline
YandexGPT-5-Lite-8B & \textbf{2.89} & 2.39 & 2.56
  & 16 & 0 & 20 \\
Mistral-7B          & 2.83 & \textbf{2.44} & 2.58
  & \textbf{18} & 0 & 16 \\
Phi-3-mini          & 2.75 & 2.28 & 2.36
  & 12 & 5 & 17 \\
GPT-2 base          & 2.37 & 2.19 & \textbf{2.66}
  & 1 & 23 & 14 \\
DistilGPT2          & 2.35 & 2.16 & 2.56
  & 9 & 4 & 21 \\
GPT-2 medium        & 2.13 & 1.89 & 2.16
  & 1 & 25 & 12 \\
\hline
\end{tabular}
\end{table}

In the direct scoring regime, YandexGPT-5-Lite-8B leads on grammar ($2.89$), Mistral-7B on coherence ($2.44$), GPT-2 base on language correctness ($2.66$). The differences between the three leading models on grammar are $2.75$--$2.89$. Models of the GPT-2 family show lower scores on all criteria ($1.89$--$2.66$); GPT-2 medium has the lowest on grammar ($2.13$) and coherence ($1.89$).

In the pairwise ranking regime, the highest win rate is for Mistral-7B (win rate $0.529$, $18$ wins and $0$ losses out of $34$ comparisons), followed by YandexGPT-5-Lite-8B ($0.444$, $16$/$0$ out of $36$), Phi-3-mini ($0.353$, $12$/5), DistilGPT2 ($0.265$, $9$/4), GPT-2 base ($0.026$, $1$/$23$) and GPT-2 medium ($0.026$, $1$/$25$). The proportion of ties for all models is $12$--$21$ out of $34$--$38$ comparisons: the judges often do not prefer either text. Win rate is reported as a descriptive characteristic: with $34$--$38$ pairs per model and $12$--$21$ ties, the statistical power of a formal test for a difference in win rate is limited.

In pairwise ranking, Mistral-7B is ahead of YandexGPT-5-Lite-8B; in direct scoring and in native-speaker evaluation (Section~\ref{s4:human}), YandexGPT-5-Lite-8B is ahead of Mistral-7B. A possible explanation is the sensitivity of pairwise comparison to length and style: in the absence of length control, pairs of Mistral-7B versus YandexGPT-5-Lite-8B may be biased towards longer texts. This is a hypothesis, not an established fact: the sample of pairs was not controlled for length, and an experiment with length control was not conducted.

The results of granular decomposition (morphology, lexicon, syntax) are not aggregated into a single metric in the main text and are not used for cross-model comparison. Spearman correlation between granular and direct criteria on the common 126 generations is $\rho = 0.60$ for the pair morphology--grammar ($p < 0.001$), $\rho = 0.22$ for lexicon--language ($p = 0.014$) and $\rho = 0.22$ for syntax--coherence ($p = 0.014$). The moderate correlation of morphology with grammar is expected: both metrics capture inflection. The weak correlation of lexicon with language fluency and of syntax with coherence indicates that the granular criteria capture aspects of quality not fully covered by direct scoring.

\subsection{Validation by Native Speakers of Bashkir}\label{s4:human}

On a random sample of 100 texts (23/23/23 for the GPT-2 family models, 11 for Mistral-7B and YandexGPT-5-Lite-8B, 9 for Phi-3-mini; the prompts and source of texts are the same as in §3.5 and Section~\ref{s4:llm_judge}), three native speakers of Bashkir assigned scores on three criteria (grammar, coherence, language) on a scale of 1--5. The unit of analysis is the text: the scores of the three annotators for each text were averaged, and all subsequent statistical comparisons (Kruskal--Wallis, Mann--Whitney~U) were conducted on the means per text. The subsamples of texts for different models do not overlap, which excludes paired comparison of models on the same texts; all pairwise comparisons were conducted as independent (Mann--Whitney U), not paired (Wilcoxon signed-rank). Annotator agreement was $\alpha = 0.772$ (grammar), $0.694$ (coherence) and $0.770$ (language correctness) --- Krippendorff's $\alpha$, an ordinal metric; all three values are above the threshold of $0.667$ for acceptable agreement \cite{Krippendorff1980ContentAA}. The mean scores by model are given in Table~\ref{tab:human}.

\begin{table}[ht]
\centering
\caption{{Scores by native speakers of Bashkir (mean over three annotators, scale 1--5, 95\% bootstrap CI)}}
\label{tab:human}
\begin{tabular}{lccccc}
\hline
Model & $n$ & Grammar & Coherence & Language & Mean \\
\hline
YandexGPT-5-Lite-8B & 11
  & \textbf{2.94 [2.70, 3.15]} & \textbf{3.09 [2.76, 3.42]} & \textbf{3.82 [3.61, 3.97]}
  & \textbf{3.28} \\
Mistral-7B          & 11
  & 2.12 [1.97, 2.30] & 2.18 [1.94, 2.48] & 3.24 [3.00, 3.48]
  & 2.52 \\
Phi-3-mini          & 9
  & 1.81 [1.56, 2.07] & 1.81 [1.48, 2.11] & 3.04 [2.85, 3.22]
  & 2.22 \\
DistilGPT2          & 23
  & 1.48 [1.28, 1.70] & 1.39 [1.22, 1.58] & 2.39 [2.16, 2.62]
  & 1.75 \\
GPT-2 base          & 23
  & 1.39 [1.20, 1.59] & 1.30 [1.14, 1.51] & 2.33 [2.16, 2.52]
  & 1.68 \\
GPT-2 medium        & 23
  & 1.16 [1.04, 1.30] & 1.22 [1.10, 1.35] & 1.70 [1.45, 1.94]
  & 1.36 \\
\hline
\end{tabular}
\end{table}

YandexGPT-5-Lite-8B leads on all three criteria (grammar $2.94$, coherence $3.09$, language correctness $3.82$), followed by Mistral-7B ($2.12$, $2.18$, $3.24$) and Phi-3-mini ($1.81$, $1.81$, $3.04$). Models of the GPT-2 family are substantially inferior ($1.16$--$2.62$), and GPT-2 medium shows the worst results on all three criteria (mean $1.36$), despite having the largest number of parameters in the group ($355$M versus $82$M and $124$M).

The differences between the six models are statistically significant for all three criteria (Kruskal--Wallis test: $H_{\text{grammar}} = 50.81$, $p = 9.4 \times 10^{-10}$; $H_{\text{coherence}} = 49.58$, $p = 1.7 \times 10^{-9}$; $H_{\text{language}} = 61.18$, $p = 6.9 \times 10^{-12}$; Table~\ref{tab:human_tests}). The effect sizes $\varepsilon^2$ are $0.47$--$0.60$ and correspond to large effects.

\begin{table}[ht]
\centering
\caption{{Kruskal--Wallis (omnibus) and number of significant pairwise comparisons (Mann--Whitney, Holm correction, family $= 15$ pairs per criterion). Subsamples of texts for different models do not overlap; pairwise comparisons are independent}}
\label{tab:human_tests}
\begin{tabular}{lcccc}
\hline
Criterion & $H$ & $p$ & $\varepsilon^2$ & Significant pairs \\
\hline
Grammar     & 50.81 & $9.4 \times 10^{-10}$ & 0.487 & 9/15 \\
Coherence   & 49.58 & $1.7 \times 10^{-9}$  & 0.474 & 9/15 \\
Language    & 61.18 & $6.9 \times 10^{-12}$ & 0.598 & 13/15 \\
\hline
\end{tabular}
\end{table}

After Holm--Bonferroni correction, 9 out of 15 pairwise comparisons were significant for grammar and coherence, and 13 out of 15 for language correctness. YandexGPT-5-Lite-8B significantly outperforms all three models of the GPT-2 family on all three criteria with effect size rank-biserial $r = +0.96$--$+0.99$ ($p_{\text{Holm}} < 0.001$ for all 9 comparisons). Mistral-7B also significantly outperforms all three models of the GPT-2 family ($p_{\text{Holm}} < 0.05$ for all 9 comparisons; $p_{\text{Holm}} < 0.01$ for 8 out of 9 --- the exception is the grammar of DistilGPT2 with $p_{\text{Holm}} = 1.3 \times 10^{-2}$), effect size $r = +0.64$--$+0.96$.

The pairwise comparison of YandexGPT-5-Lite-8B with Mistral-7B is significant on all three criteria in favour of YandexGPT-5-Lite-8B: $\Delta_{\text{grammar}} = +0.82$, $r = +0.86$, $p_{\text{Holm}} = 4.3 \times 10^{-3}$; $\Delta_{\text{coherence}} = +0.91$, $r = +0.76$, $p_{\text{Holm}} = 1.6 \times 10^{-2}$; $\Delta_{\text{language}} = +0.58$, $r = +0.66$, $p_{\text{Holm}} = 2.4 \times 10^{-2}$. The differences between Mistral-7B and Phi-3-mini do not reach significance after correction for multiple comparisons (all three criteria, $p_{\text{Holm}} > 0.05$: $0.27$, $1.00$ and $0.80$ for grammar, coherence and language respectively), which is explained by the small sample size ($n = 11$ and $n = 9$); this result does not allow the two medium-scale models to be ordered by native-speaker evaluations.

The Spearman correlation between LLM judge scores (direct scoring) and native-speaker scores was $\rho = 0.462$ for grammar ($p < 0.001$, $n = 100$) and $\rho = 0.334$ for coherence ($p < 0.001$), but did not reach significance for language correctness ($\rho = 0.162$, $p = 0.11$). The latter indicates that existing multilingual LLM judges are insufficiently sensitive to Bashkir language fluency proper; for this criterion, native-speaker evaluation remains necessary.

A comparison of the means of LLM judges and native speakers reveals a systematic difference: LLM judges give scores in the range $1.89$--$2.89$ (width $1.00$), whereas native-speaker scores cover the range $1.16$--$3.82$ (width $2.66$, $2.7$ times wider). In particular, LLM judges underestimate the superiority of YandexGPT-5-Lite-8B over the GPT-2 family on grammar: by LLM judges the gap is $2.89$ versus $2.13$--$2.37$ ($\Delta \approx 0.5$--$0.8$), by native speakers --- $2.94$ versus $1.16$--$1.48$ ($\Delta \approx 1.5$--$1.8$). Similarly, on the language criterion, LLM judges place GPT-2 base ($2.66$) above YandexGPT-5-Lite-8B ($2.56$), whereas native speakers do the opposite ($3.82$ versus $2.33$). This discrepancy should be taken into account when using LLM-as-Judge for low-resource languages: LLM judges compress the scale and in some cases change the ranking of models.

\subsection{Tokeniser Analysis}\label{s4:tokenizer}

The tokeniser analysis was conducted on the Bashkir test sample of the corpus with control on English and Turkish test sets. Fertility rate --- the average number of subtokens per word --- is given in Table~\ref{tab:fertility}. Point estimates are reported; confidence intervals are not constructed (justification in Section~\ref{s6:tokenizer}).

\begin{table}[ht]
\centering
\caption{{Fertility rate (subtokens per word) on Bashkir, English and Turkish test sets}}
\label{tab:fertility}
\begin{tabular}{lccc}
\hline
Model & Fertility (ba) & Fertility (en) & Fertility (tr) \\
\hline
GPT-2 family (DistilGPT2, base, medium) & \textbf{7.80} & 1.36 & 3.58 \\
Phi-2 (2.7B) & \textbf{7.80} & 1.36 & 3.58 \\
Mistral-7B-v0.3 & 4.81 & 1.32 & 3.75 \\
Qwen2.5-7B & 4.23 & 1.37 & 3.00 \\
YandexGPT-5-Lite-8B & 3.80 & 1.26 & 3.22 \\
T-lite-2.1 & \textbf{2.92} & 1.37 & 3.03 \\
\hline
\end{tabular}
\end{table}

In Bashkir, the fertility rate varies by a factor of $2.67$ --- from $2.92$ (T-lite-2.1) to $7.80$ (GPT-2 family, Phi-2). All three models of the GPT-2 family use a single BPE vocabulary; Phi-2 inherits the same vocabulary. YandexGPT-5-Lite-8B ($3.80$) and Mistral-7B ($4.81$) use vocabularies with coverage of Cyrillic and Turkic morphology.

Control languages give a different picture. In English, the fertility rate for all six models lies in the range $1.26$--$1.37$ (spread $1.09$ times). In Turkish --- $3.00$--$3.75$ (spread $1.25$ times). The spread in Bashkir ($2.67$) exceeds the spread in Turkish by $2.14$ times. The discrepancy is explained neither by agglutinative morphology (Turkish is agglutinative, but variability is low) nor by non-standard script (Cyrillic is present in the vocabularies of all models): the source is the low-resource status of Bashkir.

The ratio fertility$_{\text{ba}}$/fertility$_{\text{tr}}$ measures the penalty for moving from a medium-resource to a low-resource language under fixed agglutinative morphology. The greatest penalty is for the GPT-2 family and Phi-2 ($7.80/3.58 = 2.18$ times): on moving from Turkish to Bashkir, more than half of the tokens are lost. For Mistral-7B the penalty is $1.28$ times, for YandexGPT-5-Lite-8B --- $1.18$ times. T-lite-2.1 handles Bashkir no worse than Turkish ($2.92/3.03 = 0.96$).

The relationship of fertility to downstream quality is traced on three models for which downstream experiments were conducted. On the binary task, the GPT-2 family (fertility $7.80$, ratio $2.18$) shows macro-F1 $0.699$; Mistral-7B ($4.81$, $1.28$) --- $0.709$--$0.721$; YandexGPT-5-Lite-8B ($3.80$, $1.18$) --- $0.736$--$0.748$. On the multi-class task --- $0.122$, $0.282$--$0.298$, $0.332$--$0.368$ respectively. The Bashkir/Turkish fertility ratio orders these three models identically to downstream quality.

Morphological Boundary Accuracy (MorphAcc) and OOV rate are given in Table~\ref{tab:morph}.

\begin{table}[ht]
\centering
\caption{{Morphological Boundary Accuracy and OOV rate on the Bashkir test sample. MorphAcc --- the proportion of subtokens whose right boundary coincides with a morpheme boundary; OOV rate --- the proportion of word forms split into individual bytes or characters}}
\label{tab:morph}
\begin{tabular}{lcc}
\hline
Model & MorphAcc & OOV rate \\
\hline
Mistral-7B-v0.3 & 0.093 & 0.065 \\
YandexGPT-5-Lite-8B & 0.086 & 0.037 \\
GPT-2 family (DistilGPT2, base, medium) & 0.057 & 0.043 \\
Phi-2 (2.7B) & 0.057 & 0.043 \\
Qwen2.5-7B & 0.047 & 0.017 \\
T-lite-2.1 & 0.047 & 0.017 \\
\hline
\end{tabular}
\end{table}

MorphAcc in Bashkir remains low for all models: maximum $0.093$ (Mistral-7B), followed by YandexGPT-5-Lite-8B $0.086$, GPT-2 family and Phi-2 $0.057$, Qwen2.5-7B and T-lite-2.1 $0.047$. The maximum MorphAcc corresponds to models with Cyrillic vocabularies. OOV rate is also low: $0.017$--$0.065$ --- the proportion of words split into individual characters does not exceed $6.5\,\%$ for any model.

MorphAcc and OOV rate distinguish models more weakly than fertility: the range of MorphAcc ($0.047$--$0.093$) and OOV ($0.017$--$0.065$) is an order of magnitude narrower than the range of fertility ($2.92$--$7.80$). Among the three tokenisation metrics, only fertility directly orders downstream quality.

The DeepSeek-7B model is excluded from Tables~\ref{tab:fertility} and~\ref{tab:morph}: a fertility rate $0.331 < 1$ is physically impossible for a word-level tokeniser (the number of subtokens cannot be less than the number of words) and indicates an artefact of input text normalisation; the MorphAcc value $= 0.000$ for DeepSeek-7B is a consequence of the same artefact.

\subsection{One-Way Estimates of Contribution}\label{s4:variance}

To estimate the relative contribution of architecture, adaptation method and random seed, a one-way analysis of variance was applied to each group. The results for key metrics are given in Table~\ref{tab:variance}. The factors ``model'' and ``method'' are confounded in the current configuration: models of the GPT-2 family (three architectures) were trained only with full fine-tuning, and generative models of $3.8$--$8$B (three architectures) only with QLoRA. Therefore $\eta^2_{\text{method}}$ should be interpreted as the effect of the group ``small Full~FT versus large QLoRA'', and not as the pure effect of the adaptation method; separating these effects requires a crossed experimental design (Full~FT and QLoRA on the same architectures).

\begin{table}[ht]
\centering
\caption{{One-way estimates ($\eta^2$) of contribution for key metrics}}
\label{tab:variance}
\begin{tabular}{lcccc}
\hline
Metric & $\eta^2_{\text{model}}$ & $\eta^2_{\text{method}}$
  & $\eta^2_{\text{seed}}$ & $n$ \\
\hline
PPL & 0.54 & \textbf{0.73} & 0.13 & 15 \\
BPB & 0.36 & \textbf{0.72} & 0.05 & 15 \\
Target token ratio & \textbf{0.81} & 0.27 & 0.08 & 9 \\
\hline
\end{tabular}
\end{table}

For PPL, the adaptation method explains $73\,\%$ of the variance ($\eta^2 = 0.73$, $p = 0.002$), architecture --- $54\,\%$ ($\eta^2 = 0.54$, $p = 0.16$, not significant), seed --- $13\,\%$ ($\eta^2 = 0.13$, $p = 0.18$, not significant). For BPB the picture is analogous: method $72\,\%$ ($p = 0.002$), architecture $36\,\%$ ($p = 0.46$), seed $5\,\%$ ($p = 0.43$). The opposite pattern is observed for the target token ratio: $81\,\%$ of the variance is explained by architecture ($p < 0.001$), method contributes $27\,\%$ ($p = 0.15$, not significant), seed $8\,\%$ ($p = 0.47$, not significant).

Such a picture of one-way estimates indicates that the adaptation strategy dominates in explaining perplexity and BPB, whereas architecture dominates in explaining language alignment (the target token ratio). This is consistent with the hypothesis that language alignment is determined predominantly by the composition of pretraining and the tokeniser, whereas perplexity is determined by the adaptation strategy.

It should be noted that with $n = 15$ observations for PPL/BPB and $n = 9$ for the target token ratio, the analysis of variance has limited power; the results should be interpreted as preliminary. In particular, for PPL the contribution of architecture ($\eta^2 = 0.54$) does not reach the significance level ($p = 0.16$) with this sample, although the effect size is large. Final conclusions require an expansion of the number of configurations (primarily the addition of further random seeds), which constitutes a direction for further work.

\section{Discussion}\label{s5}

\subsection{Factors Behind the Superiority of YandexGPT-5-Lite-8B}\label{s5:yandex}

YandexGPT-5-Lite-8B outperforms the other models across the set of downstream metrics. A robust advantage, statistically confirmed, is established on multi-class classification by \textbf{accuracy} ($0.606$ versus $0.547$ for Mistral-7B and $0.506$ for Phi-3-mini; McNemar $p_{\text{Holm}} < 0.001$ for both pairs; Table~\ref{tab:multiclass_stats}); the point estimates of macro-F1 ($0.368$ versus $0.292$ and $0.119$) and weighted-F1 ($0.570$ versus $0.510$ and $0.430$) are consistent with this order. Superiority in BPB is also statistically confirmed ($0.462$ versus $0.549$ and $0.704$; Table~\ref{tab:lm_stats}, $p_{\text{Holm}} < 0.001$) and in native-speaker evaluations when compared with Mistral-7B ($3.28$ versus $2.52$; $p_{\text{Holm}} < 0.05$ on all three criteria; Table~\ref{tab:human}). The difference with Phi-3-mini in native-speaker evaluations ($2.22$) is pointwise in favour of YandexGPT-5-Lite-8B, but in the current sample ($n = 11$ and $n = 9$, non-overlapping texts) does not reach the significance level (Section~\ref{s4:human}). In zero-shot ICL, a pointwise superiority is observed ($0.580$--$0.679$ versus $0.360$--$0.482$ and $0.400$--$0.513$ macro-F1; Table~\ref{tab:icl}); this metric is interpreted descriptively, without a formal statistical test.

On binary classification, the statistical analysis is limited to two groups of comparisons (Table~\ref{tab:binary_stats}): the QLoRA configuration without CPT at $r=8$ and the encoder-only baseline (mBERT, XLM-R). Within the QLoRA group at $r=8$, the pointwise advantage of YandexGPT-5-Lite-8B over Mistral-7B and Phi-3-mini does not reach the significance level (Yandex vs Mistral $p_{\text{Holm}} = 0.183$, Yandex vs Phi-3 $p_{\text{Holm}} = 0.084$). The configurations at $r=16$ are presented in Table~\ref{tab:binary} as point estimates of macro-F1; their formal statistical comparison is beyond the scope of the present analysis and requires an expansion of the number of seeds (Section~\ref{s6:seed}). Among the metrics suitable for cross-model comparison, YandexGPT-5-Lite-8B is pointwise inferior to Mistral-7B only in the pairwise win rate of LLM judges ($0.444$ versus $0.529$, Table~\ref{tab:llm_judge}); this metric is interpreted descriptively, without a formal test, and the possible explanation through the absence of length control in the pairs (Section~\ref{s5:judge}) remains a hypothesis. PPL is not among the metrics for cross-model comparison: by PPL, YandexGPT-5-Lite-8B is inferior to Mistral-7B ($3.29$ versus $2.90$, Table~\ref{tab:lm}), but PPL is unsuitable for cross-model comparison with different tokenisers (Section~\ref{s6:tokenizer}).

Two reasons for the superiority of YandexGPT-5-Lite-8B follow from the analysis of tokenisers and pretraining. The first is the Cyrillic tokeniser with a lower fertility rate: $3.80$ subtokens per Bashkir word versus $7.80$ for the GPT-2 family ($2.05$ times, Table~\ref{tab:fertility}). A smaller number of subtokens per word form means greater character coverage by each token, and the model uses the limited context length ($128$ tokens in the experiments) more efficiently. The second reason is Russian-language pretraining: the Russian language uses Cyrillic and possesses agglutinative features, which gives the model partial morphological priors for Bashkir. An indirect argument in favour of this interpretation is the results of YandexGPT-5-Lite-8B in zero-shot ICL: macro-F1 $0.657$--$0.679$ in Russian and English (in Bashkir $0.580$, Table~\ref{tab:icl}), which is expected for a model with a Cyrillic tokeniser and closely related morphology.

The comparison with the GPT-2 family captures both reasons simultaneously. YandexGPT-5-Lite-8B (8B, fertility $3.80$) outperforms GPT-2 medium (355M, fertility $7.80$) by $1.92$ points on the mean of native-speaker evaluations ($3.28$ versus $1.36$) with a $22$-fold difference in the number of parameters; the fertility of Yandex is lower by a factor of $2.05$. Increasing the number of parameters with an unchanged tokeniser gives no gain: GPT-2 medium (355M) is inferior to DistilGPT2 (82M) in native-speaker evaluations ($1.36$ versus $1.75$) with the same BPE vocabulary. The comparison with Phi-3-mini is less informative for the fertility hypothesis: Phi-3-mini (3.8B) is inferior to YandexGPT-5-Lite-8B (8B) by $1.06$ points ($2.22$ versus $3.28$), but the fertility of Phi-3-mini was not measured in the present work, and the gap cannot be attributed directly to fertility. The observed picture is consistent with the hypothesis that the tokeniser and the language composition of pretraining contribute more to adaptation quality than model size in the range $3.8$--$8$B.

\subsection{The Selective Effect of Continual Pretraining}\label{s5:cpt}

The effect of continual pretraining on the Bashkir corpus (10\,000 documents, 1 epoch, QLoRA with 4-bit NF4 quantisation, $r=16$, $\text{lr}=10^{-5}$) differs across three metrics: perplexity, binary and multi-class classification. None of the metrics gives a single ranking of models by the benefit of CPT, and this discrepancy is a substantive result of the work.

\textit{Perplexity.} By PPL, CPT is positive for Phi-3-mini ($\Delta$PPL $= +0.58$, $p_{\text{Holm}} < 0.001$, Cliff's $\delta = +0.83$) and YandexGPT-5-Lite-8B ($\Delta$PPL $= +0.33$, $p_{\text{Holm}} < 0.001$, Cliff's $\delta = +0.72$) and significantly negative for Mistral-7B ($\Delta$PPL $= -1.15$, $p_{\text{Holm}} < 0.001$, Cliff's $\delta = -0.67$; Table~\ref{tab:cpt}). The direction of the effect on PPL differs across models: CPT improves language modelling for Phi-3-mini and YandexGPT-5-Lite-8B and worsens it for Mistral-7B. For the two models with a positive effect, the gain is small in absolute magnitude ($0.33$--$0.58$ PPL points), which indicates a limited benefit of CPT in the current configuration: a larger gain would require either more data (10\,000 documents --- $14\,\%$ of the full corpus of 71\,567 documents), or more epochs, or a higher learning rate. The observed selectivity is consistent with the hypothesis about the role of the volume of English-language pretraining: Mistral-7B was trained on a substantially larger English-language corpus and, presumably, possesses more stable representations which the conservative learning rate ($10^{-5}$) does not shift towards Bashkir in one epoch; for Phi-3-mini and YandexGPT-5-Lite-8B the same shift gives a positive effect. In absolute magnitude, $\Delta$PPL for Mistral-7B ($1.15$ points) is greater than for Phi-3-mini and YandexGPT-5-Lite-8B ($0.58$ and $0.33$), but with the opposite sign.

\textit{Binary classification.} On the binary task (Section~\ref{s4:binary}), the picture is different: CPT improves macro-F1 for Mistral-7B ($0.681 \to 0.721$, $+4.0$ points) and Phi-3-mini ($0.674 \to 0.701$, $+2.7$ points) and does not change the result for YandexGPT-5-Lite-8B ($0.748$ in both configurations). The greatest gain here is observed for Mistral-7B --- the very model for which CPT significantly worsened perplexity. Thus, the hypothesis ``CPT helps models with a smaller volume of multilingual pretraining'', which holds for PPL, is not confirmed on binary classification: CPT helps precisely Mistral-7B with the largest English-language pretraining.

\textit{Multi-class classification.} On the multi-class task (Section~\ref{s4:multiclass}), the effect of CPT is heterogeneous. When compared at $r=16$, CPT gives a macro-F1 gain for YandexGPT-5-Lite-8B ($0.332 \to 0.366$, $+0.034$) and Phi-3-mini ($0.153 \to 0.166$, $+0.013$), but worsens it for Mistral-7B ($-0.015$; see footnote~\ref{fn:mistral-multiclass-delta}). Here CPT is positive for the same two models as on perplexity, and negative for Mistral-7B, which is consistent with the hypothesis about the role of the volume of pretraining. A possible explanation for the negative effect for Mistral-7B is the strong class imbalance ($47.5\,\%$ of documents belong to one category), under which the signal from CPT is diluted by the dominant class; for YandexGPT-5-Lite-8B and Phi-3-mini, by contrast, CPT gives a measurable gain.

\textit{Summary.} As shown in Section~\ref{s4:cpt}, the benefit of CPT does not reduce to a single ranking of models: the sign of the effect differs both across models and across metrics, and on binary classification CPT gives the greatest gain precisely for Mistral-7B --- the model for which CPT significantly worsened perplexity. A possible explanation for the discrepancy is the different sensitivity of perplexity and downstream quality to the shift of representations under CPT: improvement in language modelling does not necessarily translate into improvement in classification and vice versa. A definitive conclusion about the benefit of CPT requires separating the effects of CPT and downstream training (for example, evaluating classification only after the CPT adapter, without QLoRA), which constitutes a direction for further work, as well as systematic variation of the learning rate and the volume of CPT data.

\subsection{Divergence Between LLM-as-Judge Regimes and Native Speakers}\label{s5:judge}

A comparison of three sources of generation quality evaluation --- LLM judges in the direct scoring regime, LLM judges in the pairwise ranking regime and native speakers of Bashkir --- reveals systematic discrepancies that should be taken into account when interpreting the results.

The first discrepancy is between direct scoring and pairwise ranking. In direct scoring, YandexGPT-5-Lite-8B leads on grammar ($2.89$ versus $2.83$ for Mistral-7B), while in pairwise ranking it is inferior ($0.444$ versus $0.529$, Table~\ref{tab:llm_judge}). At the same time, native speakers rate YandexGPT-5-Lite-8B significantly higher than Mistral-7B on all three criteria ($\Delta_{\text{grammar}} = +0.82$, $\Delta_{\text{coherence}} = +0.91$, $\Delta_{\text{language}} = +0.58$, $p_{\text{Holm}} < 0.05$ for all three, Section~\ref{s4:human}). Thus, the pairwise ranking of LLM judges gives a ranking opposite to direct scoring and native-speaker evaluations. A possible cause is the sensitivity of pairwise comparison to length and style: as shown in Section~\ref{s4:llm_judge}, in the absence of length control, pairs may be biased towards longer texts; this is a hypothesis, not an established fact. For low-resource languages this means that the pairwise ranking of LLM judges without length control should not be used as the main signal; the interpretation of the observed difference ($0.444$ vs $0.529$) remains open.

The second discrepancy is between direct scoring by LLM judges and native-speaker evaluations. The means of the LLM judges lie in the range $1.89$--$2.89$ (width $1.00$), whereas the means of native speakers lie in the range $1.16$--$3.82$ (width $2.66$, $2.7$ times wider). LLM judges compress the scale towards the middle and underestimate the gap between the best and worst models: by LLM judges, YandexGPT-5-Lite-8B outperforms GPT-2 medium on grammar by $0.76$ points ($2.89$ versus $2.13$), by native speakers --- by $1.78$ points ($2.94$ versus $1.16$). Similarly, on the language criterion, LLM judges place GPT-2 base ($2.66$) above YandexGPT-5-Lite-8B ($2.56$), whereas native speakers do the opposite ($2.33$ versus $3.82$). This indicates that multilingual LLM judges (pollux-judge-7b, M-Prometheus-7B, saiga\_llama3\_8b), trained predominantly on Russian and English, are insufficiently sensitive to Bashkir language fluency proper.

The third discrepancy is the statistical non-significance of the correlation between LLM judges and native speakers on the language criterion ($\rho = 0.162$, $p = 0.11$, Section~\ref{s4:human}). At the same time, on grammar ($\rho = 0.462$) and coherence ($\rho = 0.334$) the correlation is significant ($p < 0.001$). This means that LLM judges reproduce human evaluations when assessing formal-grammatical aspects, but not when assessing language correctness as such. For low-resource languages, where there are no available LLM judges trained on the target language, native-speaker evaluation remains necessary for the language criterion; direct scoring by LLM judges can be used as an additional signal on grammar and coherence.

\subsection{Tokenisation as a Source of Digital Inequality}\label{s5:tokenizer}

The relationship between the fertility rate of the tokeniser and adaptation quality is traced in three metrics. The fertility rate in Bashkir varies from $2.92$ (T-lite-2.1) to $7.80$ (the GPT-2 family and Phi-2), that is, by a factor of $2.67$ (Table~\ref{tab:fertility}). All three models of the GPT-2 family use a single BPE vocabulary with fertility $7.80$; Phi-2 inherits the same vocabulary. YandexGPT-5-Lite-8B ($3.80$) and Mistral-7B ($4.81$) use vocabularies covering Cyrillic and Turkic morphology.

The relationship with quality manifests itself in three metrics. By BPB, models of the GPT-2 family show $0.784$--$0.917$, whereas YandexGPT-5-Lite-8B --- $0.462$, Mistral-7B --- $0.549$ (Table~\ref{tab:lm}): with fertility $7.80$, representing the same text requires more tokens, which complicates the modelling of long dependencies. In native-speaker evaluations, models of the GPT-2 family receive $1.16$--$1.48$ on grammar versus $2.94$ for YandexGPT-5-Lite-8B (Table~\ref{tab:human}). In pairwise ranking, GPT-2 base and GPT-2 medium win $1$ out of $38$ comparisons each, whereas YandexGPT-5-Lite-8B and Mistral-7B lose none (Table~\ref{tab:llm_judge}).

The relationship is not deterministic. T-lite-2.1 has the lowest fertility ($2.92$), but did not participate in the downstream experiments; Qwen2.5-7B ($4.23$) also did not participate (Section~\ref{s6:scope}). Phi-3-mini participated in the downstream experiments, but its fertility rate was not measured in the present work: the tokeniser of Phi-3-mini differs from that of Phi-2, and the value $7.80$ is not applicable to it. The correlation is established on five models with measured fertility and downstream results: YandexGPT-5-Lite-8B, Mistral-7B, GPT-2 base, GPT-2 medium, DistilGPT2.

Control on Turkish confirms the interpretation. Turkish is an agglutinative language with Latin script and medium-resource status; in Turkish, the fertility rate for all six models of the table lies in the range $3.00$--$3.75$ (spread $1.25$ times), whereas in Bashkir --- $2.92$--$7.80$ (spread $2.67$ times). In English (high-resource) the spread is lower: $1.26$--$1.37$ ($1.09$ times). The discrepancy is consistent with the hypothesis that the source is the low-resource status of Bashkir, and not agglutinative morphology or non-standard script as such. The ratio fertility$_{\text{ba}}$/fertility$_{\text{tr}}$, which measures the penalty for low-resource status under fixed agglutinative morphology, orders the three models identically to downstream quality on the binary task: GPT-2 ($2.18$, macro-F1 $0.699$) $<$ Mistral-7B ($1.28$, $0.721$) $<$ YandexGPT-5-Lite-8B ($1.18$, $0.748$).

Increasing the number of parameters with an unchanged tokeniser does not solve the problem of digital inequality. GPT-2 medium (355M) is inferior to DistilGPT2 (82M) in native-speaker evaluations (mean $1.36$ versus $1.75$) with the same BPE vocabulary with fertility $7.80$. By contrast, YandexGPT-5-Lite-8B (8B, fertility $3.80$) outperforms GPT-2 medium (355M, fertility $7.80$) by $1.92$ points ($3.28$ versus $1.36$) with a $22$-fold difference in size: the fertility of Yandex is lower by a factor of $2.05$, which explains part of the gap. Phi-3-mini (3.8B) is inferior to YandexGPT-5-Lite-8B (8B) by $1.06$ points ($2.22$ versus $3.28$); the fertility of Phi-3-mini was not measured, so the gap cannot be attributed directly to fertility. For low-resource languages with agglutinative morphology and non-standard script, adapting the tokeniser --- extending the vocabulary with morphemes of the target language or training BPE from scratch on the target corpus --- appears to be a more promising direction than increasing the number of parameters with a fixed vocabulary.

\subsection{Practical Recommendations}\label{s5:practical}

From the set of results, four practical recommendations can be formulated for adapting LLMs to low-resource agglutinative languages with Cyrillic script.

\textit{Choice of base model.} Under a limited computation budget, preference should be given to models with a Cyrillic tokeniser and Russian-language pretraining (in the present work --- YandexGPT-5-Lite-8B), even if they are inferior in the number of parameters. Of the six metrics of cross-model comparison, YandexGPT-5-Lite-8B outperforms both competitor models (Mistral-7B and Phi-3-mini) in five: (1) multiclass accuracy --- statistically confirmed, $p_{\text{Holm}} < 0.001$; (2) BPB --- statistically confirmed, $p_{\text{Holm}} < 0.001$; (3) human eval (Yandex vs Mistral) --- statistically confirmed, $p_{\text{Holm}} < 0.05$; (4) zero-shot ICL macro-F1 --- pointwise superiority, no formal test conducted; (5) binary accuracy at $r=8$ --- pointwise superiority, not statistically confirmed ($p_{\text{Holm}} \geq 0.084$, Table~\ref{tab:binary_stats}); (6) pairwise win rate of LLM judges --- pointwise inferior to Mistral-7B ($0.444$ versus $0.529$), no significance test conducted. Thus, in four metrics out of six the superiority is either statistically confirmed or pointwise without a formal test; in one (binary accuracy) it is not statistically confirmed; in one (pairwise win rate) it is pointwise inferior. The model T-lite-2.1 has the lowest fertility rate in our analysis ($2.92$, Section~\ref{s4:tokenizer}), but did not participate in the downstream experiments; its adaptation quality remains unknown and constitutes a direction for further work.

\textit{Adaptation method.} QLoRA at $r=8$ gives macro-F1 $0.736$ on binary classification for YandexGPT-5-Lite-8B (Table~\ref{tab:binary}); at $r=16$ the same model reaches $0.748$ --- equally in configurations both without CPT and with CPT. Both results outperform full fine-tuning of the encoder-only baseline (mBERT $0.650$, XLM-R $0.653$) by $8.3$--$9.8$ points. For classification tasks, QLoRA on models of $3.8$--$8$B is pointwise preferable to full fine-tuning of encoder-only models; this comparison simultaneously changes the adaptation method, size and architecture (Section~\ref{s6:scope}) and is interpreted as an empirical observation for the selected configurations.

\textit{Continual pretraining.} CPT on $10\,000$ documents gives a PPL gain of $0.33$--$0.58$ points for Phi-3-mini and YandexGPT-5-Lite-8B ($p_{\text{Holm}} < 0.001$) and significantly worsens PPL for Mistral-7B ($\Delta = -1.15$, $p_{\text{Holm}} < 0.001$; Table~\ref{tab:cpt}). If the base model is Mistral-7B and the target metric is perplexity, CPT in the current configuration is not justified; with a downstream orientation towards binary classification, CPT for Mistral-7B gives the greatest gain ($+4.0$ points of macro-F1). For Phi-3-mini and YandexGPT-5-Lite-8B, CPT is recommended as a preliminary stage.

\textit{Quality evaluation.} For low-resource languages, direct scoring by LLM judges on the criteria of grammar and coherence reproduces native-speaker evaluations with moderate correlation ($\rho = 0.33$--$0.46$), but not on the criterion of language correctness ($\rho = 0.16$, $p = 0.11$, Section~\ref{s4:human}). Pairwise ranking without length control gives a ranking opposite to direct scoring and native-speaker evaluations (Section~\ref{s4:llm_judge}); a possible cause is the absence of length control (Section~\ref{s5:judge}). The pairwise regime without length control should not be used as the main signal. Native-speaker evaluation remains necessary for definitive conclusions about generation quality in the Bashkir language.

\section{Limitations}\label{s6}

The present study has a number of limitations that determine the scope of applicability of the conclusions and outline directions for further work. They are formulated below in descending order of significance for the interpretation of the results.

\subsection{A Single Random Seed for PEFT Experiments}\label{s6:seed}

For the QLoRA and CPT experiments (Sections~\ref{s4:binary}, \ref{s4:multiclass}, \ref{s4:cpt}), a single random seed ($42$) was used; full fine-tuning (Section~\ref{s4:lm}) was conducted with two seeds ($42$, $123$). This limits the statistical power of comparisons between PEFT configurations: the standard deviation of metrics across seeds has not been estimated, and conclusions about the superiority of one QLoRA configuration over another are formulated on the basis of point estimates and bootstrap confidence intervals over the test sample, rather than on cross-seed variability. Bootstrap intervals over the test sample (Tables~\ref{tab:binary_stats}, \ref{tab:multiclass_stats}) provide an estimate of statistical uncertainty at the level of individual examples, but not at the level of model initialisation. To confirm differences between QLoRA configurations that are close in quality (for example, YandexGPT-5-Lite-8B $0.736$ versus Mistral-7B $0.709$ in Table~\ref{tab:binary}; McNemar $p_{\text{Holm}} = 0.183$, Table~\ref{tab:binary_stats}), an expansion of the number of seeds to three or more is required.

\subsection{Multi-Class Classification}\label{s6:multiclass}

On the multi-class classification task (19 categories, $47.5\,\%$ of documents in the dominant class), no configuration exceeded macro-F1 $0.37$ (Table~\ref{tab:multiclass}). XLM-R did not outperform the majority baseline (accuracy $0.470$ versus $0.470$, McNemar $p = 1.000$); mBERT outperforms the majority baseline statistically significantly but by a small margin (accuracy $0.498$ versus $0.470$, $\Delta = +0.028$ $[+0.010, +0.047]$, $p_{\text{Holm}} = 0.006$). Both results indicate the collapse of encoder models into the dominant class under standard cross-entropy without weighting. The task remains open: solving it requires a weighted loss function, resampling or a hierarchical classification scheme. The results of the QLoRA models on the multi-class task ($0.119$--$0.368$ macro-F1) are interpreted as a lower bound on quality.

\subsection{Sample and Agreement in Native-Speaker Evaluation}\label{s6:human}

Validation by native speakers of Bashkir (Section~\ref{s4:human}) was conducted on 100 texts with three annotators. The subsamples of texts for different models do not overlap (23/23/23/11/11/9), so pairwise comparisons of models were conducted as independent (Mann--Whitney U), not paired (Wilcoxon signed-rank); this reduces power compared with a paired design on the same texts. Krippendorff's agreement lies in the range $\alpha = 0.694$--$0.772$, which exceeds the threshold of $0.667$ for acceptable agreement \cite{Krippendorff1980ContentAA}, but does not reach $0.80$ for high agreement. The sample of 100 texts is unbalanced across models ($n = 23$ for the three models of the GPT-2 family, $n = 11$ for Mistral-7B and YandexGPT-5-Lite-8B, $n = 9$ for Phi-3-mini), which limits the power of pairwise comparisons between models close in quality. In particular, the difference between Mistral-7B ($2.12$) and Phi-3-mini ($1.81$) on grammar does not reach significance after Holm correction ($p_{\text{Holm}} > 0.05$), although the pointwise difference is $0.31$ points. A definitive ordering of medium-scale models requires expanding the annotated sample to $\geq 30$ texts per model.

\subsection{Reproducibility of Splits and Possible Intersections}\label{s6:splits}

The splitting of all four datasets (corpus, binary, multiclass, multilabel) was performed with a fixed seed~$=42$ in an 80/10/10 proportion using stratified splitting by target columns; the stratification columns are described in Section~\ref{s3:datasets}. The final split sizes are: binary --- 13\,595/1\,698/1\,701, multiclass --- 14\,318/1\,790/1\,789, multilabel --- 17\,829/2\,265/2\,224, corpus --- 54\,784/6\,848/6\,848.

The binary, multiclass and multilabel datasets are formed from the news sources of the same corpus, which creates a theoretical risk of intersection between the CPT corpus and the test splits of the downstream tasks. The splitting procedure eliminates this risk: the test splits of the downstream tasks are formed first, their texts are fixed by hash, and documents matching them are excluded from the corpus before splitting. Of the 71\,567 documents of the corpus, 3\,087 (4.31\,\%) were excluded, and the remaining 68\,480 were split as 54\,784/6\,848/6\,848. A control check by text hash confirms zero intersection of the training part of the corpus with the test splits of all three downstream tasks (0 out of 1\,701, 0 out of 1\,789, 0 out of 2\,224).

\subsection{Divergence Between LLM-as-Judge and Native Speakers}\label{s6:judge}

The correlation of direct scoring by LLM judges with native-speaker evaluations is significant for grammar ($\rho = 0.462$, $p < 0.001$) and coherence ($\rho = 0.334$, $p < 0.001$), but does not reach significance for language correctness ($\rho = 0.162$, $p = 0.11$, Section~\ref{s4:human}). LLM judges systematically compress the scale (range of means $1.89$--$2.89$ versus $1.16$--$3.82$ for native speakers) and in some cases change the ranking of models: on the \textit{language} criterion, LLM judges place GPT-2 base above YandexGPT-5-Lite-8B, whereas native speakers do the opposite. This limits the use of LLM-as-Judge as the main metric for a low-resource language: judges trained predominantly on Russian and English reproduce native-speaker evaluations on formal-grammatical aspects, but are insufficiently sensitive to Bashkir language fluency proper. In the present work, LLM-as-Judge is considered as an additional signal, and not as a replacement for native-speaker evaluation; the divergence between direct scoring and pairwise ranking is discussed in Section~\ref{s5:judge}. Training a Bashkir-language LLM judge constitutes a direction for further work (Section~\ref{s6:future}).

\subsection{Scope of Tokeniser Analysis}\label{s6:tokenizer}

The tokeniser analysis (Section~\ref{s4:tokenizer}) includes three metrics: fertility rate in Bashkir, English and Turkish; Morphological Boundary Accuracy; OOV rate. MorphAcc remains low for all models ($0.047$--$0.093$), which limits its discriminative ability; the relative order by MorphAcc is consistent with the order by downstream quality, but the difference between models by MorphAcc is smaller than by fertility. The relationship of fertility to adaptation quality is correlational on a sample of five models with measured fertility and downstream results (YandexGPT-5-Lite-8B, Mistral-7B, GPT-2 base, GPT-2 medium, DistilGPT2); establishing a causal relationship requires experiments with controlled vocabulary modification. The DeepSeek-7B model is excluded from the fertility and MorphAcc analysis because of a fertility rate $0.331 < 1$, which indicates an artefact of input text normalisation.

For fertility rate, the present work reports only point estimates, without confidence intervals. The point estimate was computed as the ratio of the total number of subtokens to the total number of words on the test sample ($\sum \text{tokens} / \sum \text{words}$), whereas bootstrap resampling was built on per-document ratios ($\overline{\text{tokens}_i / \text{words}_i}$). By Jensen's inequality, the mean of ratios does not coincide with the ratio of sums; therefore, the standard percentile-bootstrap CI does not cover the point estimate, and such an interval is incorrect as an estimate of uncertainty for $\sum \text{tokens} / \sum \text{words}$. A more correct procedure --- recalculating the statistic $\sum \text{tokens}/\sum \text{words}$ on each bootstrap sample of documents --- has not been implemented. We report only point estimates of fertility and do not use them for statistical inferences about differences between models: downstream metrics and the exact McNemar test are used as the main instrument for cross-model comparisons. Bootstrap intervals for fertility are not published in the present work: the corresponding columns are excluded from the final table.

Language modelling (Section~\ref{s4:lm}) was conducted with a maximum sequence length of 128 tokens. For documents longer than this threshold, perplexity is computed on truncated sequences, which underestimates PPL for models with a high fertility rate: with fertility $7.80$ (the GPT-2 family), a test sequence of 128 tokens covers $\sim$16 Bashkir words, with fertility $3.80$ (YandexGPT-5-Lite-8B) --- $\sim$34 words. This is precisely why GPT-2 medium shows the lowest PPL ($2.33$) with the worst BPB ($0.784$): PPL is computed on short truncations, where predicting the next token is easier, whereas BPB is normalised by the number of bytes. However, even BPB does not eliminate the discrepancy completely: with the same limit of 128 tokens, models with different fertility process different fragments of the source text (with fertility $7.80$ --- the first $\sim$16 words, with $3.80$ --- the first $\sim$34 words), and a fully comparable experiment requires either a fixed length in bytes/characters or evaluation on full documents ($512$ or $1024$ tokens) with subsequent normalisation by bytes. BPB is used as the main metric for cross-model comparison; PPL values are given for completeness and are not interpreted as an indicator of language modelling quality on full documents. Controlled experiments with full document length constitute a direction for further work.

The evaluation of catastrophic forgetting (Section~\ref{s4:lm}) was conducted on the EN ($n = 200$, WikiText-2) and RU ($n = 200$, Wikipedia) test sets with only aggregated PPL and BPB preserved, but without per-example NLL. This limits the possibility of bootstrap estimation of confidence intervals and analysis of the error distribution on individual examples: the observed differences in forgetting between the three models of the GPT-2 family (BPB $1.83$--$2.08\times$ with $n = 200$) lie within statistical error and are not interpreted as significant. Forgetting was not evaluated for the QLoRA configurations, since the QLoRA experiments used a single seed and did not preserve per-example NLL for EN/RU; a direct comparison of forgetting between Full FT and QLoRA remains a direction for further work. Freezing the base weights under QLoRA is a necessary but not a sufficient condition for preserving behaviour: the adapter can change the output on EN/RU even with frozen weights, so the observed Bashkir ratio of $0.59$--$0.60$ in Table~\ref{tab:lm} is interpreted as a descriptive characteristic rather than as proven absence of forgetting.

\subsection{Scope of Model Applicability and Excluded Configurations}\label{s6:scope}

The experiments are limited to models in the range $82$M--$8$B parameters and to the Google Colab Pro computational environment (NVIDIA L4 GPU, $24$~GB). Models above $14$B parameters were not tested. Full fine-tuning was conducted for models of the GPT-2 family (DistilGPT2 $82$M, GPT-2 base $124$M, GPT-2 medium $355$M) and for the encoder-only baselines (mBERT $178$M, XLM-R $278$M); for generative models above $2.7$B parameters, full fine-tuning in the current configuration (24~GB GPU) is technically possible only up to $\sim$2.7B parameters and was not performed in the present work. The conclusions about the superiority of QLoRA over full fine-tuning (Section~\ref{s4:binary}) refer to the comparison of QLoRA on generative models of $7$--$8$B with full fine-tuning of small generative ($82$M--$355$M) and encoder-only ($178$--$278$M) models; this comparison simultaneously changes the adaptation method, size and architecture, and with full fine-tuning of generative models of $3$--$8$B the comparison may yield a different result. Separating the effects of method, size and architecture requires a crossed experimental design (Full~FT and QLoRA on the same architectures), which was not conducted in the present work.

The DoRA \cite{liu2024dora} and QDoRA methods, as well as ranks $r=8$ for language modelling, did not enter the current analysis: they are planned as an extension of the study and require additional validation of compatibility with 4-bit quantisation in the Colab environment.

ICL (Section~\ref{s4:icl}) was conducted for five models, but for two of them (DistilGPT2 and Phi-2) the generated responses did not correspond to the required format, so these models are not included in the comparative analysis. This limitation reflects adherence to the format, not the quality of task solution: the proportion of correctly recognised responses was not analysed separately; separating the effects of format and quality requires preserving a response-recognition flag for each generated text, which constitutes a direction for further work. For Phi-2, ICL was conducted only on the binary task; on the multi-class classification task, ICL was conducted for four models (DistilGPT2, Phi-3-mini, Mistral-7B, YandexGPT-5-Lite-8B) and gave macro-F1 no higher than $0.251$ (the best result --- YandexGPT-5-Lite-8B, Russian-language instruction, $K{=}1$), which is lower than QLoRA adaptation on the same task (Section~\ref{s4:multiclass}) and is not included in the main analysis.

The Qwen2.5-7B model participates in the tokeniser analysis (Section~\ref{s4:tokenizer}), but is not included in the downstream experiments: in preliminary ICL experiments it showed unstable generation in the required format (the response was not recognised by the parser), and it is excluded from the main analysis to preserve comparability with the other models. The DeepSeek-7B model is also excluded from the tokeniser analysis because of an artefact of input text normalisation (fertility rate $0.331 < 1$, Section~\ref{s6:tokenizer}).

The multi-label dataset (22\,318 documents, 14 labels) is published as a resource, but is not included in the benchmark in the present work: the micro-F1, macro-F1 and subset accuracy metrics for it are planned as an extension, since the keyword-based labelling of the dataset (Section~\ref{s3:datasets}) requires additional validation of quality before use as a reference.

\subsection{Directions for Further Work}\label{s6:future}

The set of listed limitations determines five directions for further work. First, expanding the number of random seeds for PEFT experiments to three or more with estimation of cross-seed variability. Second, solving the multi-class classification task through a weighted loss function and comparison with hierarchical schemes. Third, expanding the annotated sample for native-speaker validation to $\geq 30$ texts per model and training a Bashkir-language LLM judge. Fourth, controlled experiments with vocabulary expansion to test the causal relationship between fertility rate and adaptation quality. Fifth, a systematic investigation of approaches to generation grounded in external sources. Pilot experiments with RAG ($n=30$ questions, QLoRA-adapted Mistral-7B as generator, multilingual-e5-large as retriever) and Multi-Agent Debate ($n=10$ questions, two agents and two rounds) are not included in the main analysis: on a sample of $n=6$ common questions, pairwise comparison revealed no significant differences (faithfulness $p_{\text{boot}} = 0.056$, answer relevance $p_{\text{boot}} = 0.912$), the absolute scores of both approaches remain below the neutral point of the scale (faithfulness $1.80$--$2.50$, answer relevance $1.63$--$2.13$ on a scale of 1--5), and the sample is insufficient for definitive conclusions. Artefacts of the pilot experiments are available in the repository; a systematic investigation of RAG and MAD in Bashkir with an increased sample and tuning of the retriever component constitutes a separate direction. These directions go beyond the scope of the present work, but follow directly from the results obtained.

\section{Conclusion}\label{s7}

The work presents an open resource package for the Bashkir language and a systematic benchmark of methods for adapting large language models to a low-resource agglutinative language with Cyrillic script. The main artefacts are the Bashkir Web Corpus (71\,567 documents, 46.9 million tokens, 16 sources), three labelled datasets for binary, multi-class and multi-label classification, as well as trained adapters for all configurations presented in Section~\ref{s4}. The experiments were conducted on eight architectures in the range $82$M--$8$B parameters and three groups of adaptation strategies: full fine-tuning (small models and encoder-only baselines), QLoRA (medium-scale generative models) and in-context learning (zero-shot and few-shot); in addition, the benefit of continual pretraining on the Bashkir corpus as a preliminary stage before QLoRA adaptation was tested.

The main empirical result on binary classification is that QLoRA on the 8B model YandexGPT-5-Lite-8B reaches macro-F1 $0.736$ at $r=8$ and $0.748$ at $r=16$ (the value $0.748$ is achieved both without CPT and with CPT), which is $8.3$--$9.8$ points higher than full fine-tuning of the encoder-only baseline (mBERT $0.650$, XLM-R $0.653$, Table~\ref{tab:binary}). In a paired comparison of accuracy of the QLoRA configurations without CPT ($0.737$ versus $0.659$ and $0.652$), the difference is significant after Holm correction for $15$ comparisons (McNemar, $p_{\text{Holm}} < 0.001$, Table~\ref{tab:binary_stats}); the statistical conclusions refer to accuracy, while macro-F1 values are given as point estimates. This comparison simultaneously changes the adaptation method, the size and the architecture of the model; it is interpreted as an empirical observation for the selected configurations, and not as a pure comparison of methods, and a comparison of QLoRA with full fine-tuning on the same architectures was not conducted in the present work (Section~\ref{s6:scope}). Within the QLoRA group at $r=8$, differences in accuracy are not significant: Yandex vs Mistral $p_{\text{Holm}} = 0.183$, Yandex vs Phi-3 $p_{\text{Holm}} = 0.084$; the pointwise advantage of YandexGPT-5-Lite-8B is not statistically confirmed.

On multi-class classification, YandexGPT-5-Lite-8B significantly outperforms Mistral-7B and Phi-3-mini in accuracy ($0.606$ versus $0.547$ and $0.506$, $p_{\text{Holm}} < 0.001$; Table~\ref{tab:multiclass_stats}); the point estimates of macro-F1 ($0.368$ versus $0.292$ and $0.119$) and weighted-F1 ($0.570$ versus $0.510$ and $0.430$) are consistent with this order; the statistical conclusions refer to accuracy. In BPB of language modelling, YandexGPT-5-Lite-8B also significantly outperforms both models ($0.462$ versus $0.549$ and $0.704$, $p_{\text{Holm}} < 0.001$; Table~\ref{tab:lm_stats}). According to native-speaker evaluations of Bashkir, YandexGPT-5-Lite-8B significantly outperforms Mistral-7B on all three criteria ($3.28$ versus $2.52$, $p_{\text{Holm}} < 0.05$, $n=11$); the difference with Phi-3-mini ($2.22$) in the current sample ($n=11$ and $n=9$, non-overlapping texts) does not reach the significance level because of the small sample size (Section~\ref{s6:human}). Zero-shot ICL shows a pointwise superiority of YandexGPT-5-Lite-8B ($0.580$--$0.679$ versus $0.360$--$0.482$ and $0.400$--$0.513$ macro-F1; Table~\ref{tab:icl}), although a formal test for this metric was not conducted. The only metric on which YandexGPT-5-Lite-8B is pointwise inferior to Mistral-7B is the pairwise win rate of LLM judges ($0.444$ versus $0.529$); a statistical test for a difference in win rate was not conducted, and the possible explanation through the absence of length control in the pairs (Section~\ref{s5:judge}) remains a hypothesis rather than an established fact. PPL is not among the metrics for cross-model comparison: by PPL, YandexGPT-5-Lite-8B is inferior to Mistral-7B ($3.29$ versus $2.90$), but PPL is unsuitable with different tokenisers (Section~\ref{s6:tokenizer}).

Continual pretraining on $10\,000$ documents of the Bashkir corpus (QLoRA with 4-bit NF4 quantisation, $r=16$, one epoch, $\text{lr}=10^{-5}$) gives a significant perplexity gain for Phi-3-mini ($\Delta = +0.58$, $p_{\text{Holm}} < 0.001$, Cliff's $\delta = +0.83$) and YandexGPT-5-Lite-8B ($\Delta = +0.33$, $p_{\text{Holm}} < 0.001$, Cliff's $\delta = +0.72$), but significantly worsens perplexity for Mistral-7B ($\Delta = -1.15$, $p_{\text{Holm}} < 0.001$, Cliff's $\delta = -0.67$; Table~\ref{tab:cpt}). The effect of CPT on downstream tasks is directed differently. On binary classification, CPT gives the greatest gain precisely for Mistral-7B ($+4.0$ points of macro-F1); on multi-class classification, CPT improves YandexGPT-5-Lite-8B ($+0.034$) and Phi-3-mini ($+0.013$), but worsens Mistral-7B ($-0.015$). For Mistral-7B, the directions of the CPT effect on LM and on binary classification are opposite: perplexity increases by $1.15$ points, while macro-F1 increases by $4.0$ points. The benefit of CPT depends both on the base model and on the task; a definitive conclusion about the benefit of CPT requires separating the effects of CPT and downstream training (for example, evaluating classification only after the CPT adapter, without QLoRA), which was not conducted in the present work.

The fertility rate of the tokeniser in Bashkir varies by a factor of $2.67$ --- from $2.92$ (T-lite-2.1) to $7.80$ (the GPT-2 family and Phi-2; Table~\ref{tab:fertility}) --- and is consistent with the observed adaptation quality on a sample of five models with measured fertility and downstream results. With fertility $7.80$, models of the GPT-2 family show BPB $0.784$--$0.917$ and native-speaker scores $1.16$--$1.48$, whereas YandexGPT-5-Lite-8B ($3.80$) --- BPB $0.462$ and mean $3.28$. Perplexity is misleading: GPT-2 medium shows the lowest PPL ($2.33$) with the worst BPB ($0.784$), which makes BPB the main metric for cross-model comparison with different tokenisers. The mechanism of this discrepancy is explained in Section~\ref{s6:tokenizer}: when test sequences are truncated to 128 tokens, models with a high fertility rate cover fewer Bashkir words per sequence, and with an unchanged limit in tokens different tokenisers process different fragments of text. Control on Turkish confirms the interpretation: the spread of fertility in Bashkir ($2.67$ times) exceeds the spread in Turkish ($1.25$ times) and English ($1.09$ times), and the ratio fertility$_{\text{ba}}$/fertility$_{\text{tr}}$ ($2.18$ for GPT-2, $1.28$ for Mistral-7B, $1.18$ for YandexGPT-5-Lite-8B) orders the three models identically to downstream quality on the binary task.

Four practical recommendations follow from the set of results. Under a limited computation budget, preference should be given to models with a Cyrillic tokeniser and Russian-language pretraining (in the present work --- YandexGPT-5-Lite-8B), even if they are inferior in the number of parameters. For classification tasks, QLoRA on models of $3.8$--$8$B is pointwise preferable to full fine-tuning of the encoder-only baseline: all QLoRA configurations significantly outperform mBERT, XLM-R and the majority baseline in accuracy on the binary task ($p_{\text{Holm}} < 0.001$, Table~\ref{tab:binary_stats}); at the same time, the comparison simultaneously changes the method, size and architecture and is interpreted as an empirical observation for the selected configurations. CPT is justified on perplexity for Phi-3-mini and YandexGPT-5-Lite-8B, but significantly worsens PPL for Mistral-7B ($p_{\text{Holm}} < 0.001$); at the same time, on binary classification CPT gives the greatest gain precisely for Mistral-7B ($+4.0$ points of macro-F1), so the choice of CPT configuration should be determined by the task. For evaluating generation quality in a low-resource language, direct scoring by LLM judges on the criteria of grammar and coherence reproduces native-speaker evaluations with moderate correlation ($\rho = 0.33$--$0.46$), but not on the criterion of language correctness ($\rho = 0.16$, $p = 0.11$); native-speaker evaluation remains necessary.

The limitations of the present work (a single random seed for PEFT experiments, the collapse of encoder-only models on the multi-class task, the imbalance of the sample in native-speaker evaluation, the correlational nature of the relationship between fertility and adaptation quality on a sample of five models, and the truncation of test sequences to 128 tokens in perplexity evaluation) determine the immediate directions for further work: expanding the number of seeds to three or more; a weighted loss function for multi-class classification; expanding the annotated sample to $\geq 30$ texts per model and training a Bashkir-language LLM judge; controlled experiments with vocabulary expansion to test the causal relationship between fertility rate and adaptation quality; and a systematic investigation of approaches to generation grounded in external sources (RAG and Multi-Agent Debate).

The artefacts of the study are described in Section~\ref{s3:reproducibility}. The resource package and benchmark are intended for further research on the adaptation of large language models to the Bashkir language and to other low-resource agglutinative languages with non-standard script.

\bibliographystyle{unsrtnat}
\bibliography{references}  

\end{document}